\documentclass[acmsmall]{acmart}

\setcopyright{rightsretained}
\acmDOI{10.1145/3649816}
\acmYear{2024}
\copyrightyear{2024}
\acmSubmissionID{oopslaa24main-p18-p}
\acmJournal{PACMPL}
\acmVolume{8}
\acmNumber{OOPSLA1}
\acmArticle{99}
\acmMonth{4}
\received{21-OCT-2023}
\received[accepted]{2024-02-24}
\startPage{1}

\setcitestyle{nosort}

\usepackage{booktabs}   
\usepackage{subcaption} 
\usepackage{xspace}
\usepackage{makecell}
\usepackage{tabularx}
\usepackage{tikz}
\usepackage{xcolor}
\usepackage{multirow}
\usepackage{array}
\usepackage{listings}
\usepackage{hhline}
\usepackage{subcaption}
\usepackage[ruled, lined, linesnumbered, commentsnumbered, longend, vlined]{algorithm2e}
\usepackage{algorithmic}
\usepackage{amsmath}
\usepackage{enumitem}
\usepackage{etoolbox}
\usepackage{adjustbox}
\usepackage{changepage}
\usepackage{siunitx}
\setlist[enumerate]{nosep}
\newcolumntype{?}{!{\vrule width 1.2pt}}
\newenvironment{psmallmatrix}
  {\left(\begin{smallmatrix}}
  {\end{smallmatrix}\right)}
\usepackage{letltxmacro}
\LetLtxMacro\orgvdots\vdots
\LetLtxMacro\orgddots\ddots

\makeatletter
\DeclareRobustCommand\vdots{%
  \mathpalette\@vdots{}%
}
\newcommand{\circled}[2][]{%
  \tikz[baseline=(char.base)]{%
    \node[shape = circle, draw, inner sep = -0.3pt]
    (char) {\phantom{\ifblank{#1}{#2}{#1}}};%
    \node at (char.center) {\makebox[0pt][c]{#2}};}}
\robustify{\circled}
\newcommand{\pgftextcircled}[1]{
    \setbox0=\hbox{#1}%
    \dimen0\wd0%
    \divide\dimen0 by 2%
    \begin{tikzpicture}[baseline=(a.base)]%
        \useasboundingbox (-\the\dimen0,0pt) rectangle (\the\dimen0,1pt);
        \node[circle,draw,outer sep=0pt,inner sep=0.1ex] (a) {#1};
    \end{tikzpicture}
}

\newcommand*{\@vdots}[2]{%
  \sbox0{$#1\cdotp\cdotp\cdotp\m@th$}%
  \sbox2{$#1.\m@th$}%
  \vbox{%
    \dimen@=\wd0 %
    \advance\dimen@ -3\ht2 %
    \kern.5\dimen@
    \dimen@=\wd2 %
    \advance\dimen@ -\ht2 %
    \dimen2=\wd0 %
    \advance\dimen2 -\dimen@
    \vbox to \dimen2{%
      \offinterlineskip
      \copy2 \vfill\copy2 \vfill\copy2 %
    }%
  }%
}
\DeclareRobustCommand\ddots{%
  \mathinner{%
    \mathpalette\@ddots{}%
    \mkern\thinmuskip
  }%
}
\newcommand*{\@ddots}[2]{%
  \sbox0{$#1\cdotp\cdotp\cdotp\m@th$}%
  \sbox2{$#1.\m@th$}%
  \vbox{%
    \dimen@=\wd0 %
    \advance\dimen@ -3\ht2 %
    \kern.5\dimen@
    \dimen@=\wd2 %
    \advance\dimen@ -\ht2 %
    \dimen2=\wd0 %
    \advance\dimen2 -\dimen@
    \vbox to \dimen2{%
      \offinterlineskip
      \hbox{$#1\mathpunct{.}\m@th$}%
      \vfill
      \hbox{$#1\mathpunct{\kern\wd2}\mathpunct{.}\m@th$}%
      \vfill
      \hbox{$#1\mathpunct{\kern\wd2}\mathpunct{\kern\wd2}\mathpunct{.}\m@th$}%
    }%
  }%
}
\newcommand{\IRName}{UniSparse\xspace} 
\newcommand{\Tree}{metadata tree\xspace} 
\newcommand{\TREE}{Metadata Tree\xspace}

\newcommand{\Map}{index map\xspace} 
\newcommand{\MAp}{Index map\xspace} 
\newcommand{\MAP}{Index Map\xspace} 

\newcommand{\rev}[1]{\textcolor{black}{#1}}

\newcommand{\newrev}[1]{\textcolor{black}{#1}}
\newcommand{\newtablerev}{\color{black}}

\newcommand{\pie}[1]{%
  \begin{tikzpicture}[baseline=-0.5ex]
    \draw (0,0) circle (1ex);\fill (1ex,0) arc (0:#1:1ex) -- (0,0) -- cycle;
    \end{tikzpicture}%
}

\makeatletter
\newcommand{\twowayarrows}[2]{%
  \mathrel{\mathop{%
    \vcenter{\offinterlineskip\m@th
      \ialign{\hfil##\hfil\cr
        \hphantom{$\scriptstyle\mspace{8mu}{#1}\mspace{8mu}$}\cr
        \rightarrowfill\cr
        \vrule height0pt width 2em\cr
        \leftarrowfill\cr
        \hphantom{$\scriptstyle\mspace{8mu}{#2}\mspace{8mu}$}\cr
        \noalign{\kern-0.3ex}
      }%
    }%
  }\limits^{#1}_{#2}}%
}
\makeatother

\newcommand{\TrimOp}{\lstinline[morekeywords={trim}]{trim}\xspace}

\newcommand{\MergeOp}{\lstinline[morekeywords={merge}]{merge}\xspace}

\newcommand{\ConvertOp}{\lstinline[morekeywords={convert}]{convert}\xspace}
\newcommand{\DecomposeOp}{\lstinline[morekeywords={decompose}]{decompose}\xspace}

\newcommand{\SumOp}{\lstinline[morekeywords={sum}]{sum}\xspace}
\newcommand{\EnumerateOp}{\lstinline[morekeywords={enumerate}]{enumerate}\xspace}
\newcommand{\ReorderOp}{\lstinline[morekeywords={reorder}]{reorder}\xspace}
\newcommand{\ScheduleOp}{\lstinline[morekeywords={schedule}]{schedule}\xspace}
\newcommand{\PartitionOp}{\lstinline[morekeywords={partition}]{partition}\xspace}
\newcommand{\PackOp}{\lstinline[morekeywords={pack}]{pack}\xspace}

\newsavebox{\mylistingbox}

\lstdefinelanguage{syntax}{
xleftmargin=2em, framexleftmargin=2em, aboveskip=1mm, belowskip=1mm,
numberstyle=\scriptsize,
tabsize=2,
columns=fullflexible,
basicstyle={\scriptsize\ttfamily},
breaklines=true,
breakatwhitespace,
escapeinside=<>,
alsoletter={\.},
identifierstyle={\color{black}},
numbers=left,
numbersep=8pt,
}

\lstdefinelanguage{emit}{
xleftmargin=2em, framexleftmargin=2em, aboveskip=1mm, belowskip=1mm,
numberstyle=\scriptsize,
tabsize=2,
columns=fullflexible,
basicstyle={\scriptsize\ttfamily},
breaklines=true,
breakatwhitespace,
escapeinside=<>,
alsoletter={\_},
identifierstyle={\color{black}},
numbers=left,
numbersep=8pt,
comment=[l]{//},
commentstyle=\color{olive}\ttfamily\bfseries,,
morekeywords = [1]{if, while, for, return, in, of, to, each},
keywordstyle = [1]\color{purple}\bfseries,
morekeywords = [2]{indexMapMatrix, tile\_union, tile\_split, trim, merge, split, fill, devectorize, vectorize, sum, enumerate, enum, reorder, schedule, pack, partition, contains, apply},
keywordstyle = [2]\color{brown},
morekeywords = [3]{step 1, step 2, step 3},
keywordstyle = [3]\color{olive}\bfseries,
morekeywords = [3]{Algorithm, Input, Output, FormatConversion, T, S, T_format, S_format},
keywordstyle = [3]\color{black}\bfseries,
}

\lstdefinelanguage{mlir}{
  alsoletter={\_, \#,\%, \=, \-, \.},
  morekeywords = [1]{tensor, f32},
  keywordstyle = [1]\color{orange},
  morekeywords = [2]{linalg, unisparse, arith},
  keywordstyle = [2]\color{violet}\bfseries,
  morekeywords = [3]{decompose, convert, generic, mulf, addf, yield},
  keywordstyle = [3]\color{violet}\bfseries,
  morekeywords = [4]{div, floordiv, mod},
  keywordstyle = [4]\color{gray},
  literate={0}{{\textcolor{purple}{0}}}{1}%
         {1}{{\textcolor{purple}{1}}}{1}%
         {2}{{\textcolor{purple}{2}}}{1}%
         {3}{{\textcolor{purple}{3}}}{1}%
         {4}{{\textcolor{purple}{4}}}{1}%
         {5}{{\textcolor{purple}{5}}}{1}%
         {6}{{\textcolor{purple}{6}}}{1}%
         {7}{{\textcolor{purple}{7}}}{1}%
         {8}{{\textcolor{purple}{8}}}{1}%
         {9}{{\textcolor{purple}{9}}}{1}
         {f32}{{\textcolor{orange}{f32}}}{1}
         {A1}{{\textcolor{black}{A1}}}{1}
         {A2}{{\textcolor{black}{A2}}}{1}
         {\%A21}{{\textcolor{olive}{\%A2}}}{1}
         {\%0}{{\textcolor{black}{\%0 }}}{1}
         {bb0}{{\textcolor{black}{bb0}}}{1}
         {d0}{{\textcolor{black}{d0}}}{1}
         {d1}{{\textcolor{black}{d1}}}{1}
         {\%1}{{\textcolor{black}{\%1 }}}{1}
         {\%3}{{\textcolor{black}{\%3 }}}{1}
         {j-i}{{j\textcolor{gray}{-}i}}{1}
         {[\#COO}{{[\textcolor{teal}{\#COO}}}{1}
         {[\#BDIA}{{[\textcolor{teal}{\#BDIA}}}{1}
         {[\#C2SR}{{\textcolor{teal}{\#C2SR}}}{1}
         {[\#BELL}{{[\textcolor{teal}{\#BELL}}}{1}
         ,
  morekeywords = [5]{\#CSR, \#COO, \#DCSR, \#LIL, \#DOK, \#BDIA, \#DIA, \#DIA\-variant, \#BCSR, \#CSB, \#ELL, \#C\2SR, \#CISR, \#CISR\-plus, \#CPSR, \#Serpens, \#BELL, \#COO\_COO, \#BDIA\_CSR, \#BELL\_COO, \#map, \#encoding, \#map, \#prim, \#spmv, \#hybrid, \#sum, \#unisparse\.encoding, \#unisparse\.hybrid, \#unisparse\.sum, \#unisparse\.map, \#unisparse\.prim},
  keywordstyle = [5]\color{teal},
  morekeywords = [6]{affine\_map, ins, outs},
  keywordstyle = [6]\color{brown},
  identifierstyle=\color{black},
  sensitive=false,
  comment=[l]{//},
  morecomment=[s]{/*}{*/},
  commentstyle=\color{olive}\ttfamily,
  stringstyle=\color{purple}\ttfamily,
  morestring=[b]',
  morestring=[b]"
}

\lstdefinelanguage{codegen}{
  alsoletter={\_, \#,\%, \=, \-, \.},
  morekeywords = [1]{range, len},
  keywordstyle = [1]\color{brown},
  morekeywords = [2]{for, while, if, in, and, ==},
  keywordstyle = [2]\color{purple}\bfseries,
  morekeywords = [5]{iterate,access\_metadata,equals,access\_value,compute},
  keywordstyle = [5]\color{black}\bfseries,
  identifierstyle=\color{black},
  sensitive=false,
  comment=[l]{//},
  morecomment=[s]{/*}{*/},
  commentstyle=\color{olive}\bfseries\ttfamily,
  stringstyle=\color{purple}\ttfamily,
  morestring=[b]',
  morestring=[b]"
}

\begin{document}


\title{\IRName: An Intermediate Language for General Sparse Format Customization}


\author{Jie Liu}
\orcid{0000-0003-1534-3500}
\affiliation{%
  \institution{Cornell University}
  \city{Ithaca}
  \country{USA}
}
\email{jl3952@cornell.edu}

\author{Zhongyuan Zhao}
\orcid{0000-0002-6637-553X}
\affiliation{%
  \institution{Cornell University}
  \city{Ithaca}
  \country{United States}
}
\email{zhozh@qti.qualcomm.com}

\author{Zijian Ding}
\orcid{0009-0000-4555-2077}
\affiliation{%
  \institution{University of California}
  \city{Los Angeles}
  \country{USA}
}
\email{bradyd@cs.ucla.edu}

\author{Benjamin Brock}
\orcid{0000-0003-1488-1622}
\affiliation{%
  \institution{Intel}
  \city{San Jose}
  \country{USA}
}
\email{benjamin.brock@intel.com}

\author{Hongbo Rong}
\orcid{0000-0002-3275-7791}
\affiliation{%
  \institution{Intel}
  \city{San Jose}
  \country{USA}
}
\email{hongbo.rong@intel.com}

\author{Zhiru Zhang}
\orcid{0000-0002-0778-0308}
\affiliation{%
  \institution{Cornell University}
  \city{Ithaca}
  \country{USA}
}
\email{zhiruz@cornell.edu}
\begin{abstract}

The ongoing trend of hardware specialization has led to a growing use of custom data formats when processing sparse workloads, which are typically memory-bound. These formats facilitate optimized software/hardware implementations by utilizing \rev{sparsity pattern-} or target-aware data structures and layouts to enhance memory access latency and bandwidth utilization. However, existing sparse tensor programming models and compilers offer little or no support for productively customizing the sparse formats. \rev{Additionally, because these frameworks represent formats using a limited set of per-dimension attributes, they lack the flexibility to accommodate numerous new variations of custom sparse data structures and layouts.}

To overcome this deficiency, we propose \IRName, an intermediate language that provides a unified abstraction for representing and customizing sparse formats. Unlike the existing attribute-based frameworks, \IRName decouples the logical representation of the sparse tensor (i.e., the data structure) from its low-level memory layout, enabling the customization of both. As a result, a rich set of format customizations can be succinctly expressed in a small set of well-defined query, mutation, and layout primitives. We also develop a compiler leveraging the MLIR infrastructure, which supports adaptive customization of formats, and automatic code generation of format conversion and compute operations for heterogeneous architectures. We demonstrate the efficacy of our approach through experiments running commonly-used sparse linear algebra operations with specialized formats on multiple different hardware targets, including an Intel CPU, an NVIDIA GPU, an AMD Xilinx FPGA, and a simulated processing-in-memory (PIM) device.

\end{abstract}



\maketitle

\section{Introduction}
As Dennard scaling ended in the mid-2000s and Moore's Law is approaching its limit, computer engineers are increasingly turning to special-purpose hardware accelerators to meet the ever-growing computational demands of emerging application domains such as graph analytics, machine learning, and robotics. At the same time, there has been an explosion in the amount of data that domain experts have to manage. 
Notably, much of this big data is sparse in nature. For example, Amazon co-purchase graphs have 400K nodes and a density of 0.002\%, and arXiv graph datasets have 100M papers and a density of 0.00002\%~\cite{hu-ogb-neurips2020,snapnets}. 
These evident trends in technology and applications are driving computing systems towards heterogeneity that can process sparse data in an efficient and high-performance manner.

Many important operations (i.e., kernels) of sparse processing are performed on sparse tensors, a generalization of sparse matrices. 
Sparse tensors are typically represented using specialized \emph{data structures} that leverage the sparsity of the tensor to reduce storage size and/or memory footprint. These data structures usually store only the non-zero elements (or non-zero blocks) of the tensor, along with their associated coordinates that are encoded in a compressed form as \emph{metadata}.
Various forms of \emph{data layouts}, such as the structure of arrays (SoA) or array of structures (AoS), can be employed to store the sparse data structure in memory. The data structure and data layout jointly determine a \emph{sparse format}. 

The metadata can be viewed as a hierarchical tree that captures the multi-dimensional coordinates of the non-zeros in a structured way. In this work, we refer to this tree as the \emph{metadata tree} and use it as a logical representation of the sparse format.
To reconstruct the original coordinates of a non-zero element, multiple indirect memory accesses are required to traverse the \Tree. Due to the input-dependent and irregular data access patterns that result from this process, sparse workloads are typically memory-bound. 

To efficiently utilize memory bandwidth, reduce memory accesses, and exploit data parallelism to boost the performance of sparse tensor computation, researchers are increasingly using custom sparse formats optimized for particular application domains and/or target hardware architectures. Examples include hybrid formats for GPUs~\cite{bell2009implementing,choi2010model,guo2016hybrid} and banked formats for
FPGAs
~\cite{hu2021graphlily,fowers2014high} and dedicated accelerators~\cite{srivastava2020matraptor}. 
While format customization can significantly improve performance, we recognize two pressing issues: i) \emph{productivity} -- it takes substantial engineering effort to design a custom sparse format and adapt the implementation of related compute operations
that must interact with the new format\newrev{, and}
ii) \emph{permutability} -- there lacks a unified abstraction that can systematically encode different variants (or permutations) of existing sparse formats to facilitate the exploration of a complex design space, where the search of  custom formats needs to account for \rev{non-zero distribution patterns of inputs, }
inherent parallelism of the dominant compute kernels, and the target hardware. 

Prior research has attempted to address the productivity challenge by using either manually optimized libraries or automatic compilers. Sparse linear/tensor algebra libraries (e.g., sparse BLAS, Intel MKL, NVIDIA cuSPARSE)
provide highly optimized target-specific sparse kernels. While library functions achieve high performance, they only support a limited set of sparse formats.
Recent efforts on sparse tensor algebra compilers such as TACO~\cite{kjolstad2017tensor,chou2018format}, COMET~\cite{tian2021high}, and SparseTIR~\cite{ye2022sparsetir}  describe tensor dimensions in attributes (e.g., \lstinline{dense} or \lstinline{compressed}),
and generate sparse tensor algebra kernels assisted by predefined code generation templates. 
This attribute-based format abstraction limits their extensibility to support new custom formats, as the finite combinations of attributes restrict the range of possible data structures, and the fixed code generation templates restrict the possible data layouts. As a result, this abstraction offers programmers no further customization opportunities for the data structures and layouts, and 
important details about the data structures and layouts necessary for identifying a specific format uniquely may be omitted.
This work proposes \emph{\IRName}, which is the first intermediate language designed for general sparse format customization. \IRName aims to (1) provide a systematic way of expressing an unlimited number of custom formats, (2) support format customization at both the logical data structure and physical layout level, while taking into account the \rev{sparsity patterns of input tensors}, compute operations, and hardware targets, and (3) automate code generation for compute operations and conversion with other formats for the newly defined formats.
With \IRName, the \Tree serves as a logical representation that can be expressed using an \emph{\Map}, a set of \emph{query} and \emph{structural mutation} operations, which we call \emph{primitives} (\S \ref{coordinate-tree}). An additional set of \emph{layout} primitives specifies how to partition and traverse the \Tree, which transforms the tree into physical memory layouts (\S \ref{layout}). The index map and primitives are the essential components of a succinct \emph{intermediate language} for specifying custom sparse formats, including but not limited to many previously proposed high-performance formats. 
Compared to the previous attribute-based approach,  the \IRName language offers a holistic representation of sparse formats using mapping functions and primitives, without presuming dimension-wise composition of separate data structures. The language decouples logical representations of sparse formats from their physical memory layouts, allowing both to be specialized. Empowered by the language, the \IRName compiler can reason formally about the correspondence between various formats and their underlying layouts, automating both code generation and format conversion. To this end, we develop a data structure and data layout inference algorithm (\S \ref{layout-inference}) that automatically determines the storage format of sparse tensors. Furthermore, we introduce a format conversion algorithm (\S \ref{format-conversion}) that enables the compiler to handle a broad range of source and destination formats, including both conventional and specialized ones. We also provide a general compute kernel generation algorithm (\S \ref{compute-codegen}) that supports custom sparse formats.

The \IRName compiler is built on top of the MLIR infrastructure~\cite{lattner2020mlir} (\S \ref{IR}). We include the compiler as an artifact for evaluation, and in Section~\ref{evaluation}, we demonstrate its efficacy in customizing formats on multiple hardware platforms, including CPUs, GPUs, FPGAs, and a simulated PIM device~\cite{devic2022pim}. It can also automate the conversion among a variety of sparse formats, resulting in significant productivity improvements. 

\section{Background and Motivation}
This section overviews common sparse formats (Figure \ref{fig:format-storage}), and prior research and their limitations to motivate our work (\S \ref{prior-work}). Sparse matrices are used as illustrative examples for simplicity, while the discussion generalizes to tensors.

\subsection{Sparse Formats}

\begin{figure*}[ht]
    \centering
    \begin{subfigure}{0.24\textwidth}
        \centering
        \includegraphics[scale=0.35]{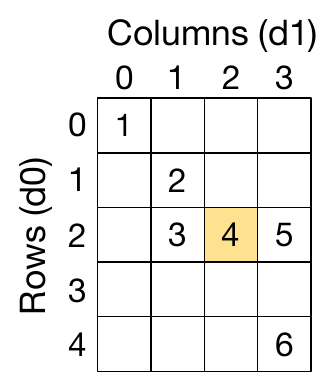}
        \caption{A 5x4 matrix $A$}
        \label{matrix}
    \end{subfigure}
    \begin{subfigure}{0.24\textwidth}
        \centering
        \includegraphics[scale=0.35]{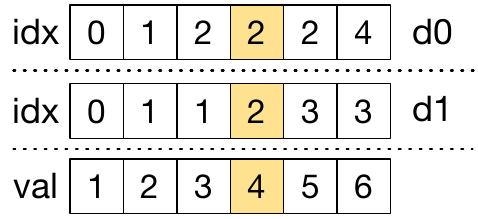}
        \caption{COO}
        \label{COO}
    \end{subfigure}
    \begin{subfigure}{0.24\textwidth}
        \centering
        \includegraphics[scale=0.35]{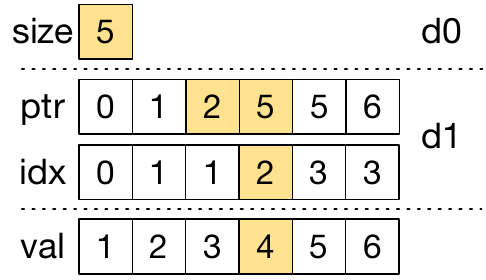}
        \caption{CSR}
        \label{CSR}
    \end{subfigure}
    \begin{subfigure}{0.24\textwidth}
        \centering
        \includegraphics[scale=0.35]{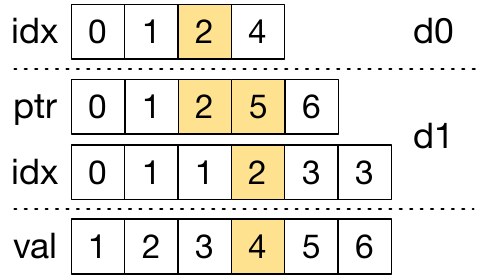}
        \caption{DCSR}
        \label{DCSR}
    \end{subfigure}
    \begin{subfigure}{0.26\textwidth}
        \centering
        \includegraphics[scale=0.35]{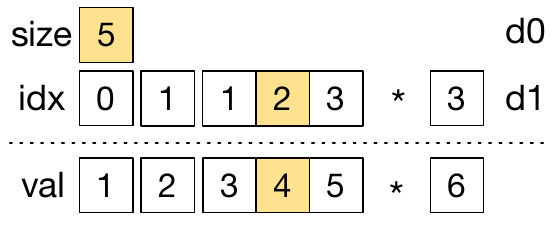}
        \caption{LIL}
        \label{LIL}
    \end{subfigure}
    \begin{subfigure}{0.24\textwidth}
        \centering
        \includegraphics[scale=0.35]{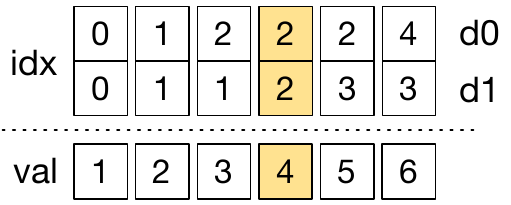}
        \caption{DOK}
        \label{DOK}
    \end{subfigure}
    \begin{subfigure}{0.22\textwidth}
        \centering
        \includegraphics[scale=0.35]{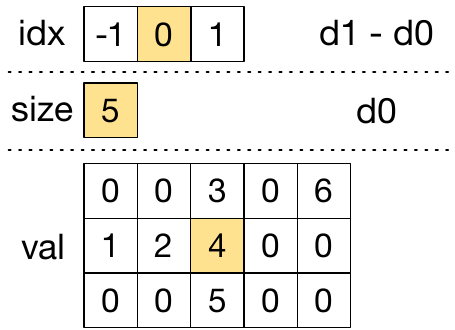}
        \caption{DIA}
        \label{DIA}
    \end{subfigure}
    \begin{subfigure}{0.22\textwidth}
        \centering
        \includegraphics[scale=0.35]{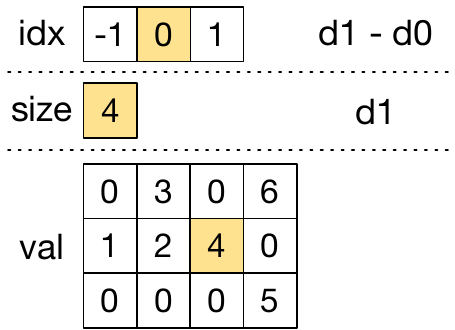}
        \caption{DIA-variant}
        \label{DIA-variant}
    \end{subfigure}
    \begin{subfigure}{0.19\textwidth}
        \centering
        \includegraphics[scale=0.35]{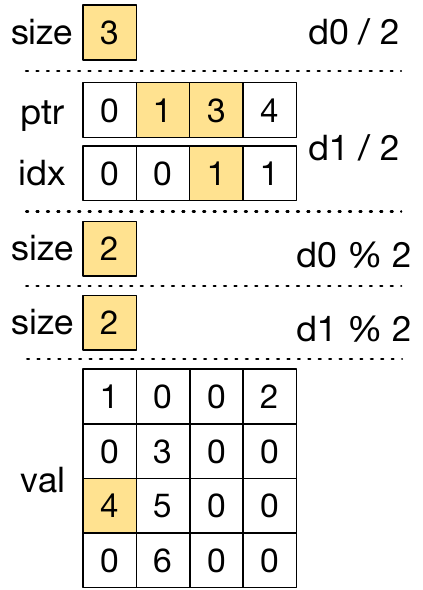}
        \caption{BCSR}
        \label{BCSR}
    \end{subfigure}
    \begin{subfigure}{0.25\textwidth}
        \centering
        \includegraphics[scale=0.35]{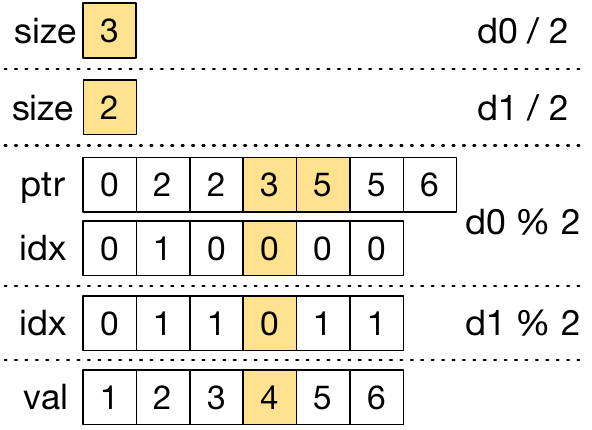}
        \caption{CSB}
        \label{CSB}
    \end{subfigure}
    \begin{subfigure}{0.22\textwidth}
        \centering
        \includegraphics[scale=0.35]{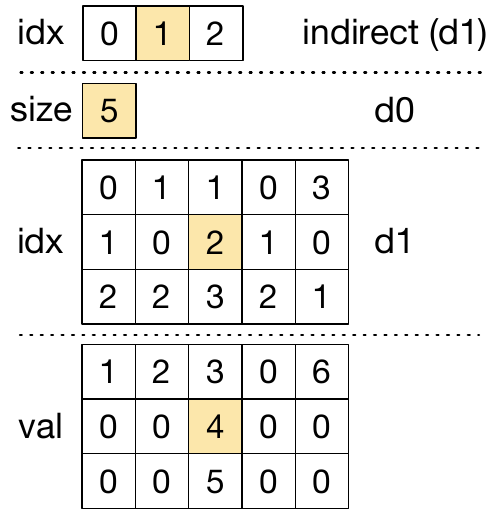}
        \caption{ELL}
        \label{ELL}
    \end{subfigure}
    \begin{subfigure}{0.27\textwidth}
        \centering
        \includegraphics[scale=0.35]{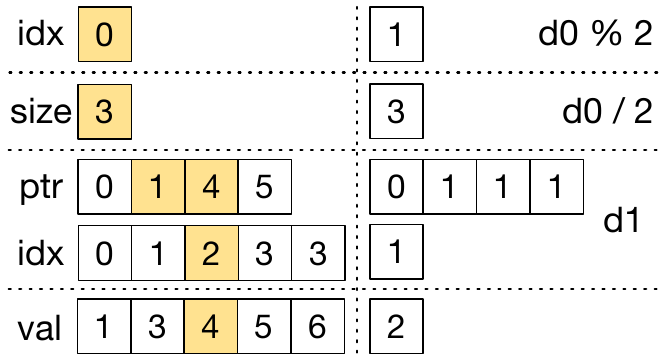}
        \caption{C\textsuperscript{2}SR (2 banks)}
        \label{C2SR}
    \end{subfigure}
    \begin{subfigure}{0.28\textwidth}
        \centering
        \includegraphics[scale=0.35]{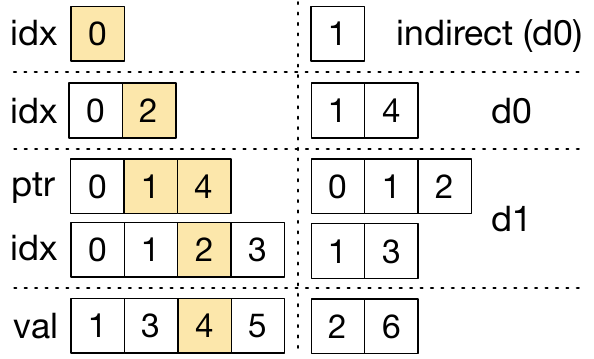}
        \caption{CISR (2 banks)}
        \label{CISR}
    \end{subfigure}
    \begin{subfigure}{0.4\textwidth}
        \centering
        \includegraphics[scale=0.35]{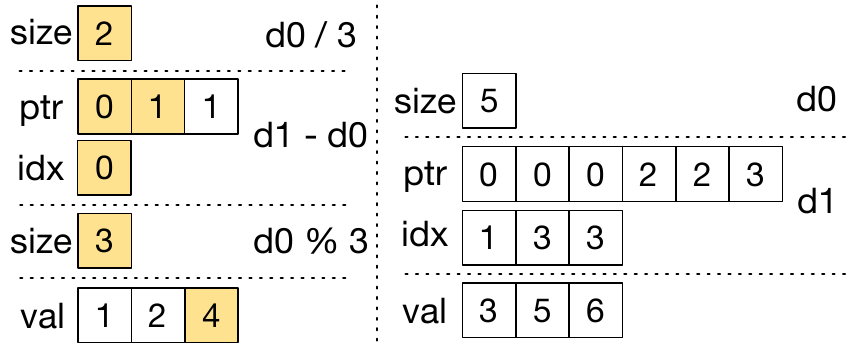}
        \caption{Hybrid BDIA/CSR}
        \label{BDIA-CSR}
    \end{subfigure}
    \begin{subfigure}{0.3\textwidth}
        \centering
        \includegraphics[scale=0.35]{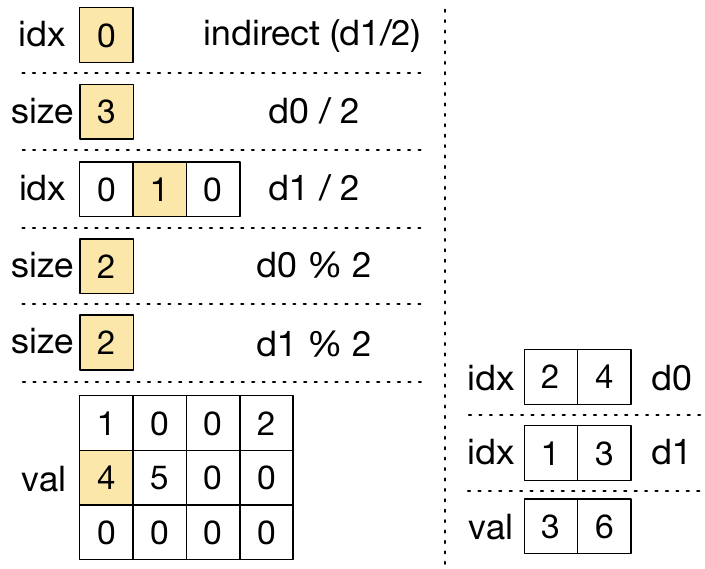}
        \caption{Hybrid BELL/COO}
        \label{BELL-COO}
    \end{subfigure}
    \vspace{-5pt}
    \caption{Different formats of a sparse matrix ($A$) --- Yellow-shaded blocks refer to the data structure of the same tensor element. The \textit{size}, \textit{ptr}, \textit{idx}, and \textit{val} labels on the left indicate the size of a dimension, a pointer array, an index array, or a value array. The index expressions on the right denote the \Map at each level. Horizontal dotted lines separate different tensor dimensions, while vertical dotted lines divide the matrix $A$ into sub-matrices. 
    }
    \label{fig:format-storage}
    \vspace{-8pt}
\end{figure*}
To improve performance and adapt to various architectures and sparsity patterns, numerous tensor formats have been developed. 


The coordinate (COO) format~\cite{bader2008efficient} (Figure \ref{COO}) stores non-zero values along with their complete coordinate information. COO is widely used as the default format for many sparse data files, such as the Matrix Market (.mtx)~\cite{boisvert1996matrix} and FROSTT (.tns)~\cite{frosttdataset} file formats. 
While COO stores row and column indices as separate arrays, the dictionary of keys (DOK) format~\cite{johansson2015sparse} (Figure \ref{DOK}) adopts a different memory layout by pairing up indices in a single array.
Sparse tensors may have multiple non-zero values per row, and many non-zero elements can share a common row index. The compressed sparse row (CSR) format (Figure \ref{CSR}) replaces these row coordinates with a compact pointer array that slices values belonging to different rows, thus saving more space than COO. The linked list (LIL) format~\cite{johansson2015sparse} (Figure \ref{LIL}) employs an alternative physical layout of the CSR format, where column indices and values are stored separately per row, eliminating the need for row pointers. The doubly-compressed sparse row (DCSR) format~\cite{buluc2008representation} (Figure \ref{DCSR}) provides further optimizations for matrices with many empty rows by storing the pointers array as a compressed list and avoiding storing pointers for empty rows in the pointers array. 

To efficiently store matrices with non-zero values clustered along the diagonals, the diagonal (DIA) format~\cite{doi:10.1137/1.9780898718003} (Figure \ref{DIA}) stores only the diagonals containing non-zero values. The traditional DIA format pads diagonals to the full size of the row dimension (\lstinline{d0} in Figure \ref{fig:format-storage}), while a variant of the DIA format (Figure \ref{DIA-variant}) pads diagonals to the full size of the column dimension (\lstinline{d1}), which is more space-efficient for matrices with fewer columns. 

Blocked formats partition a tensor into sub-tensor chunks.
For instance, a blocked variant of CSR, known as the block compressed sparse row (BCSR) format~\cite{im1998model} (Figure \ref{BCSR}), stores a compressed collection of small dense matrix blocks. In contrast, the compressed sparse block (CSB) format~\cite{Bulu2009ParallelSM} (Figure \ref{CSB}) stores a dense collection of matrix blocks in a compressed format. BCSR improves register reuse, while CSB exposes parallel execution opportunities in sparse matrix-vector multiplication (SpMV).
Another format that balances workloads for vectorized processing is the ELLPACK (ELL) format~\cite{kincaid1989itpackv} (Figure \ref{ELL}). It packs non-zero values in each row and reduces the column size to the maximum amount of non-zero values per row.

Hardware accelerators for sparse processing can benefit from  formats customized per data access patterns or memory systems.
One such format is the cyclic channel sparse row (C\textsuperscript{2}SR) format~\cite{srivastava2020matraptor} (Figure \ref{C2SR}). This format partitions a tensor's rows into sub-tensors, one sub-tensor for each memory bank, increasing memory bandwidth utilization.
Another example is the compressed interleaved sparse row (CISR) format~\cite{fowers2014high} (Figure \ref{CISR}), which 
schedules rows of the CSR format to different compute units of an accelerator, balancing the workload and eliminating memory access conflicts. 

Formats can also be customized based on the sparsity patterns of input tensors. A hybrid format represents sub-tensors with their best suitable formats independently.
One example is the hybrid blocked-DIA (BDIA) format ~\cite{fukaya2021accelerating} and CSR format (Figure \ref{BDIA-CSR}), which achieves higher performance for datasets with non-zeros clustered along diagonals on multi-thread CPUs by increasing data temporal locality in cache blocks. Another example is the hybrid blocked-ELLPACK (BELL)/COO format (Figure \ref{BELL-COO}), which combines several format optimization techniques ~\cite{guo2016hybrid, bell2009implementing,choi2010model}. 



\subsection{Prior Work on Sparse Format Abstraction}
\label{prior-work}
Early sparse tensor algebra compilers \cite{bik1993compilation,pugh1999sipr} transform dense linear algebra programs to runnable sparse code with sparsity predicates (guards), but only a few formats are supported with hard-coded format descriptions.
The idea of supporting different data structures with a format abstraction was pioneered by the Bernoulli Compiler \cite{kotlyar1997compiling}, which introduces an index hierarchy and per-dimension accessing rules to describe various sparse formats.
Another work \cite{arnold2010specifying} defines a functional language for specifying sparse matrix formats as a sequence of constructs that facilitates code verification. 

Recent years have seen a surge of research on sparse tensor compilers. TACO \cite{kjolstad2017tensor, chou2018format} represents sparse formats as per-dimension attributes and generates code for tensor algebra expressions based on the attributes.
Automated format conversion is proposed in \cite{chou2020automatic}, which uses {\Map}s and queries to specify formats in a more programmable way. The MLIR SparseTensor dialect \cite{bik2022compiler} is a recent effort that leverages TACO's code generation theories to build a sparse linear algebra compiler within the MLIR infrastructure. Other compiler frameworks, such as COMET \cite{tian2021high}, also use attribute-based format abstractions similar to TACO and target heterogeneous architectures and computational chemistry workloads. Additionally, SparseTIR \cite{ye2022sparsetir} generates high-performance sparse kernels for machine learning workloads on GPUs using hybrid formats. Table \ref{tab:prior-work} summarizes recent work on sparse tensor algebra compilers with format abstraction, categorized into two classes:
\begin{table}[ht]
\centering
\footnotesize
\vspace{-5pt}
\caption{State-of-the-art sparse tensor algebra compilers.}
\vspace{-5pt}
\sffamily 
\begin{tabularx}{1\textwidth}
{>{\centering\arraybackslash}X>{\centering\arraybackslash}c>{\centering\arraybackslash}c>{\centering\arraybackslash}c>{\centering\arraybackslash}c>{\centering\arraybackslash}c}
\hline
\textbf{Abstraction} & \textbf{Prior Work} & \thead{Custom\\{\MAP}s} & \thead{\rev{Sparsity Pattern-Aware}\\Hybrid Formats} & \thead{Backend-Aware\\Memory Layouts} & \thead{Automated \\Conversion} \\
\hline
\multirow{6}{*}{Attribute-based}  & \makecell{TACO\\ \cite{kjolstad2017tensor},\\ \cite{chou2018format}}  & \pie{0}  &\pie{0} & \pie{180} & \pie{0} \\\cline{2-6}
 &  \makecell{MLIR SparseTensor\\~\cite{bik2022compiler}} &\pie{180}  &\pie{0}& \pie{180} & \pie{180} \\\cline{2-6}
  & \makecell{COMET\\\cite{tian2021high}} & \pie{0}  &\pie{0} & \pie{180} & \pie{0} \\\cline{2-6}
   & \makecell{SparseTIR\\\cite{ye2022sparsetir}} & \pie{180}  &\pie{180} & \pie{180} & \pie{0} \\
\hline
\multirow{5}{*}{Language-based}  & \makecell{LL\\\cite{arnold2010specifying}}  & \pie{180}  &\pie{0} & \pie{180} & \pie{0} \\\cline{2-6}
& \makecell{TACO-conversion\\~\cite{chou2020automatic}}  & \pie{180}  &\pie{0} & \pie{180} & \pie{360} \\\cline{2-6}
& \makecell{\IRName\\(This Work)} & \pie{360}  &\pie{360} & \pie{360} & \pie{360} \\
\hline
\end{tabularx} 
\label{tab:prior-work}
\vspace{-5pt}
\end{table}

\textbf{Attribute-Based Format Abstraction.}
Previous research, such as TACO~\cite{kjolstad2017tensor, chou2018format}, MLIR's SparseTensor dialect~\cite{bik2022compiler}, COMET~\cite{tian2021high}, and SparseTIR~\cite{ye2022sparsetir}, uses per-dimension attributes to describe formats.
For example, in TACO, the CSR format is encoded as \lstinline|{dense, compressed}| in row and column dimensions, respectively. Similarly, SparseTIR uses \lstinline|{I = dense_fixed(I_Size, dataType), J = sparse_variable(I, ..., dataType)}| to represent the same format. 

However, limited by the finite combinations of attributes, the attribute-based approach is unable to support a vast number of new formats. For instance, while TACO can express the traditional DIA format (Figure \ref{DIA}), it cannot accommodate the DIA-variant format (Figure \ref{DIA-variant}). Additionally, TACO lacks support for custom {\Map}s, which are necessary for the CISR format (Figure \ref{CISR}). MLIR's SparseTensor dialect and SparseTIR have incorporated index expressions, which enhance expressiveness, but do not entirely overcome the limitation. Moreover, the prior approach generates compute operations with fixed iteration and access templates, resulting in non-customizable memory layouts. Additionally, different memory layouts such as SoA and AoS are not distinguished. 
Lastly, since attributes do not directly reflect the memory layouts, implementing fully automated format conversion becomes challenging. 

\textbf{Language-Based Format Abstraction.} 
Earlier research \cite{arnold2010specifying} defines a sparse format by specifying the compression process using a functional language. 
However, this approach strongly couples the specification of compute operations with the format description, leading to significant variations in user programs for the same compute kernel with different formats. 
Another work~\cite{chou2020automatic} introduces a language to assist format conversion. It supports \Map functions and blocking formats, and thus can handle formats such as the DIA-variant (Figure \ref{DIA-variant}), BCSR (Figure \ref{BCSR}), and CSB (Figure \ref{CSB}). However, it still does not support custom {\Map}s required by CISR (Figure \ref{CISR}), or discernible layouts required by LIL (Figure \ref{LIL}) and DOK (Figure \ref{DOK}). Moreover, \rev{sparsity pattern}-aware hybrid formats are not supported.

To the best of our knowledge, \IRName is the first language for format abstraction that can encode a wide range of custom formats including those with custom {\Map}s, hybrid formats aware of \rev{sparsity patterns}, and target-specific memory layouts.



\section{Overview of \IRName}
\begin{figure}[ht]
\centering
\includegraphics[width=0.63\columnwidth]{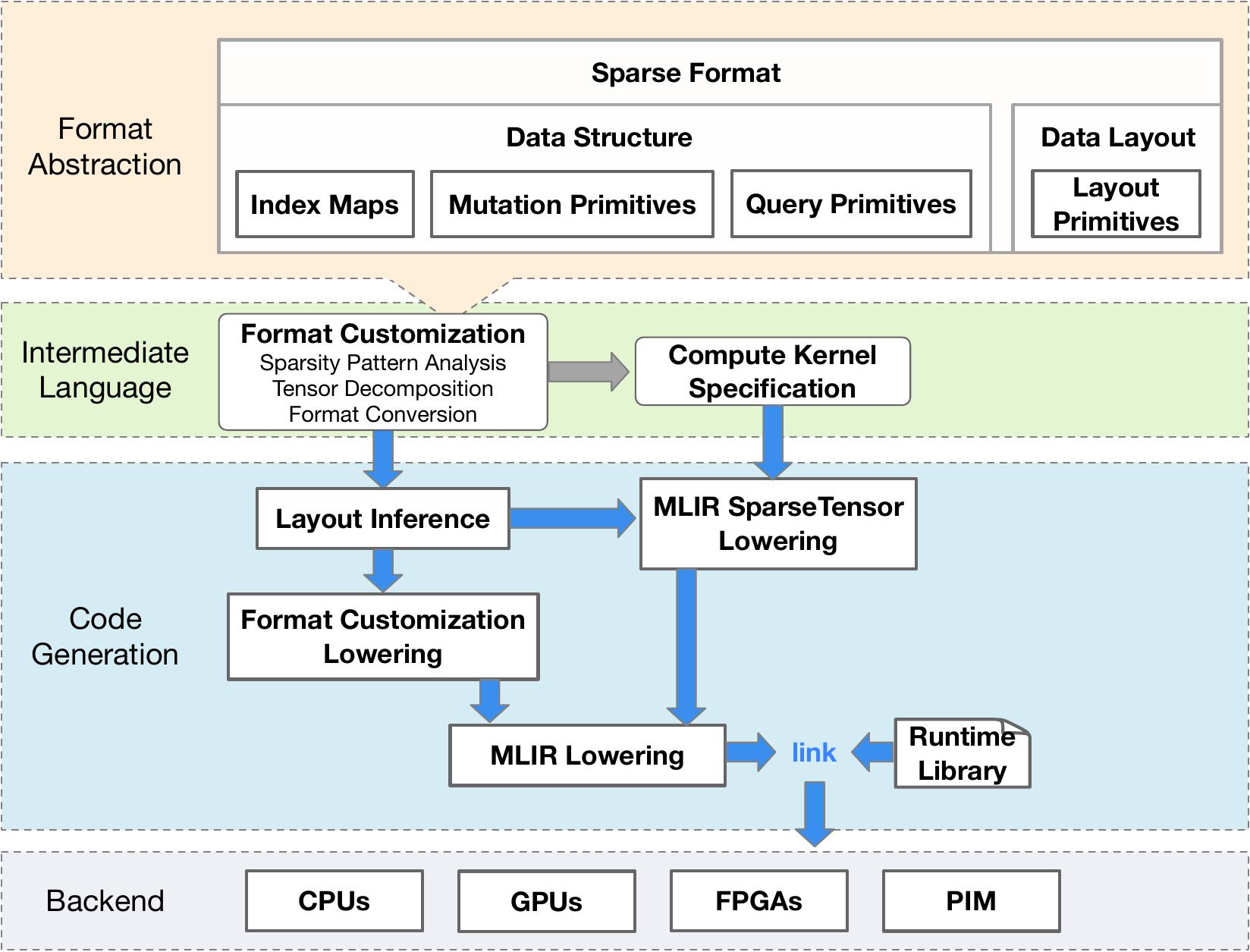} 
\vspace{-8pt}
\caption{An overview of \IRName.
}

\label{fig-overview}
\vspace{-8pt}
\end{figure}
\IRName is an intermediate language designed to formally and succinctly represent a wide range of sparse formats, along with a compiler that automates both format conversion and customization. Figure \ref{fig-overview} offers an overview of \IRName's design, with the format abstraction forming the foundation of the entire framework. We develop an intermediate language that enables users to specify both format customization and compute kernels independently. The intermediate language program can run on various backends through automated code generation.

Our approach to sparse tensor format abstraction revolves around a succinct yet expressive representation for both data structures and layouts. The data structures of a format are logically represented as a \Tree (\S \ref{coordinate-tree}), encoded by an \Map (\S \ref{coordinate-maps}) and a set of query (\S \ref{query}) and mutation primitives (\S \ref{storage-mutation}). From the \Tree, we obtain the data layouts of a format using layout primitives (\S \ref{layout}). 

During the code generation stage (\S \ref{codegen}), the \IRName compiler first infers the data structures and layouts from format encodings (\S \ref{layout-inference}). Once the formats are determined, the compiler proceeds to lower format customization and compute operations. \rev{For format conversion, the compiler automatically applies a sequence of rewrite rules to convert from the source format to the destination format (\S \ref{format-conversion}). To lower compute operations, \IRName leverages the MLIR SparseTensor dialect for its supported conventional formats. The compute kernels for custom formats are generated based on the proposed two-step algorithm (\S \ref{compute-codegen}).}
The \IRName intermediate language (\S \ref{IR}) is implemented as a new dialect in MLIR.
Our compiler can support multiple hardware targets, including CPUs, GPUs, FPGAs, and a PIM simulator~\cite{devic2022pim}. 
As shown in Figure \ref{fig-overview}, \IRName follows a design principle of decoupling format abstraction and compiler implementation at three levels: 
\begin{itemize}[leftmargin=*,noitemsep,topsep=0pt]
\item Data structures and memory layouts are expressed in a decoupled way, enabling support for a broader range of custom sparse formats.
\item The format customization and compute operations are specified independently, freeing developers from tedious format-specific implementation details when writing compute kernels.
\item The format abstraction is independent of the input language, making it portable and complementing prior attribute-based approaches. 
\end{itemize}

\begin{figure*}
    \centering
    \begin{subfigure}{0.3\textwidth}
        \centering
        \includegraphics[scale=0.3]{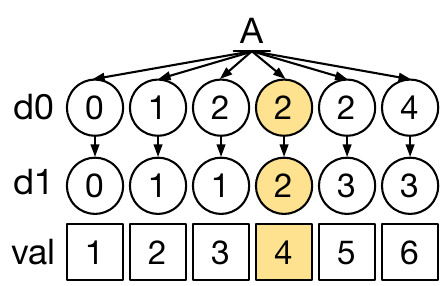}
        \caption{Source data structure: COO}
        \label{fig:illustrate-source-format}
    \end{subfigure}
    \begin{subfigure}{0.34\textwidth}
        \centering
        \includegraphics[scale=0.3]{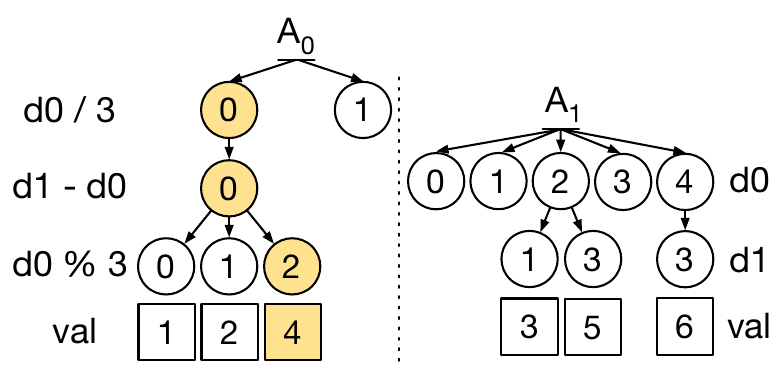}
        \caption{Target data structure: BDIA/CSR}
        \label{fig:illustrate-target-format}
    \end{subfigure}
    \begin{subfigure}[b]{0.3\textwidth}
    \begin{lstlisting}[
    basicstyle={\scriptsize\ttfamily},
    identifierstyle={\color{black}},
    numbersep=0pt,
    breaklines=true,
    xleftmargin=0em, framexleftmargin=0em, aboveskip=0mm, belowskip=0mm,
    escapeinside=<>
    ]
    // COO to BDIA
    COO_structure
    .Skew(0, 1, -1)
    .TileSplit(0, 3).Sort()
    .Fill(0).Merge(0)
    // COO to CSR
    COO_structure
    .Fill(0).Merge(0) 
    \end{lstlisting}
    \caption{Convert COO to BDIA/CSR.}
    \label{fig:illustrate-emitted-operators}
    \end{subfigure}
    \begin{subfigure}{\textwidth}
        \centering
\begin{minipage}{\columnwidth}
\begin{lstlisting}[
basicstyle={\scriptsize\ttfamily},
identifierstyle={\color{black}},
language={mlir},
numbersep=6pt,
numbers=left,
xleftmargin=2em, framexleftmargin=2em, aboveskip=1mm, belowskip=1mm,
escapeinside=**
]
// Format abstraction
#COO = #unisparse.encoding<{  *\label{spmv-format-abstraction}*
        idx_map = #unisparse.map<(d0,d1)->(d0,d1)>, mutation = #unisparse.prim<trim(0,1)> }> 
#CSR = #unisparse.encoding<{ 
        idx_map = #unisparse.map<(d0,d1)->(d0,d1)>, mutation = #unisparse.prim<merge(0), trim(1,1)> }>
#BDIA = #unisparse.encoding<{ 
        idx_map = #unisparse.map<(d0,d1)->(d0/3,d1-d0,d0%3)>, mutation = #unisparse.prim<merge(0), trim(1,1)> }>
#COO_COO = #unisparse.hybrid<{ fmats = [#COO, #COO] }>  
#BDIA_CSR = #unisparse.hybrid<{ fmats = [#BDIA, #CSR] }> *\label{spmv-format-abstraction-end}*
// Format pre-processing
#sum = #unisparse.sum< groupBy (d0,d1) -> (d0/3,d1-d0), with val ne 0 -> 1 | otherwise -> 0 > *\label{spmv-sum-query}*
%A1 = unisparse.decompose (%in_A, %thld) { query = #sum }: tensor<?x?xf32, #COO_COO> *\label{spmv-decompose}*
%A2 = unisparse.convert (%A1): tensor<?x?xf32, #BDIA_CSR> *\label{spmv-convert}*
// Compute operation
#spmv = { indexing_maps = [ *\label{spmv-signature}*
  affine_map<(d0,d1)->(d0,d1)>, // for argument %A21 *\label{spmv-map-s}*
  affine_map<(d0,d1)->(d1)>,     // for argument %in_X
  affine_map<(d0,d1)->(d0)>],  // for argument %out_Y *\label{spmv-map-e}*
  iterator_types = ["parallel", "reduction"] }  *\label{spmv-signature-end}*
%0 = linalg.generic #spmv ins(%A2, %in_X : tensor<?x?xf32, #BDIA_CSR>, tensor<?xf32>) *\label{spmv-linalg-ins}*
    outs(%out_Y: tensor<?xf32>) { *\label{spmv-linalg-generic}*
    ^bb0(%A_val: f32, %X_val: f32, %Y_val: f32):
      %1 = arith.mulf %A_val, %X_val : f32 *\label{spmv-compute-start}*
      %o = arith.addf %Y_val, %1 : f32 *\label{spmv-compute-end}*
      linalg.yield %o : f32
  } -> tensor<?xf32>  *\label{spmv-linalg-generic-end}*
\end{lstlisting}
\end{minipage}
\caption{A \IRName program of the SpMV kernel with the format of the input tensor converted from COO to hybrid BDIA/CSR. The blocking size of the BDIA format is 3. File input operations are omitted. The \lstinline{decompose} operation in Line \ref{spmv-decompose} divides the tensor into two sub-tensors adaptively by embedding a \SumOp primitive that queries the sparsity \rev{pattern} of the tensor data. A \lstinline[morekeywords={convert}]{convert} operation in Line \ref{spmv-convert} translates the source into the target format. The SpMV kernel is specified using a \lstinline[alsoletter={\.},morekeywords={linalg\.generic}]{linalg\.generic} operation in Line~\ref{spmv-linalg-generic} - \ref{spmv-linalg-generic-end}.}
\label{fig:illustrate-unisparse-code}
\end{subfigure}

\vspace{-6pt}
\caption{An illustration of \IRName.}
\vspace{-10pt}
\label{fig:illustration}
\end{figure*}

\textbf{An Illustrative Example.}
Figure~\ref{fig:illustration} illustrates our approach using the SpMV kernel with an input matrix (Figure \ref{matrix}), originally in the COO format (Figure \ref{fig:illustrate-source-format}) and then converted to the hybrid BDIA/CSR format (Figure \ref{fig:illustrate-target-format}). 
Although we will cover the formal syntax and technical details of the \IRName language in later sections, the purpose of this example is to provide an intuitive understanding of how we address the following questions: (1) The sparsity \rev{patterns of the matrices are} input-dependent. How to decompose the matrix according to the \rev{non-zero distribution patterns}? (2) How to express the source and target format in a general way? Here we use COO and BDIA/CSR for illustration purposes only, while in practice, other formats can be used. 
(3) For productivity, conversion between the formats should be automated. How could our compiler figure out the memory layouts of the formats, and automatically convert one to the other? (4) For generality, it is desirable to write a compute kernel only once but the kernel works with a matrix in any format after conversion. How to decouple the compute kernel from a specific sparse format?            

In Figure \ref{fig:illustrate-unisparse-code}, Line \ref{spmv-format-abstraction}-\ref{spmv-format-abstraction-end} specify the source and target formats. The sparsity \rev{pattern} of the input is queried (Line \ref{spmv-sum-query}), and the result is used to decompose the source format into two parts, both in the original source format COO (Line \ref{spmv-decompose}). Then, the decomposed source format is converted to the target format (Line \ref{spmv-convert}). This target format is a parameter to the compute kernel (Line \ref{spmv-signature}-\ref{spmv-linalg-generic-end}). In other words, the compute kernel and the format of its input matrix are completely decoupled.
According to the above \IRName program, the \IRName compiler automatically infers the memory layout of the source format (Figure \ref{COO}), and emits a sequence of internal operators (Figure \ref{fig:illustrate-emitted-operators}), which, once executed, result in the memory layout of the target format (Figure \ref{BDIA-CSR}).

\section{Tensor Format Abstraction}

This section describes the proposed format abstraction that supports custom data structures and layouts for sparse tensors. We first describe the expression of sparse data structures in Section \ref{coordinate-tree}, which is logically represented as a \Tree with {\Map}s, query, and mutation primitives. The \Tree representation and {\Map}s are first introduced in TACO~\cite{kjolstad2017tensor, chou2020automatic}. 
This work extends {\Map}s to express a diverse range of custom formats. We further propose a new encoding method that holistically expresses data structures in primitives, without presuming per-dimension data structure separation. Section \ref{layout} shows how memory layouts are determined by applying layout primitives to the \Tree, allowing formats to be expressed and customized at the physical layout level, which is also our new contribution. 

\subsection{\TREE}
\label{coordinate-tree}
\begin{figure*}[ht]
\centering
\subfloat[CSR or LIL]{\>\includegraphics[scale=0.32]{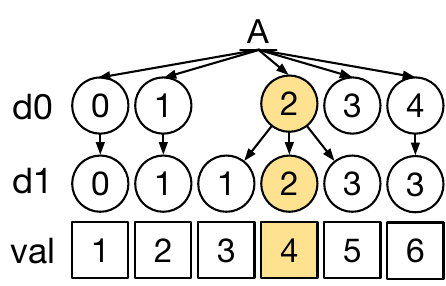}\>%
\label{CSR-tree}}
\subfloat[DCSR]{\>\includegraphics[scale=0.32]{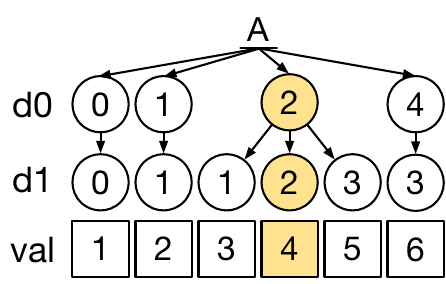}%
\label{DCSR-tree}}
\subfloat[C\textsuperscript{2}SR]{\>\includegraphics[scale=0.32]{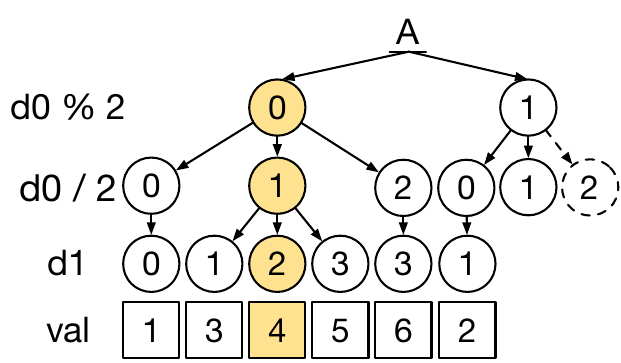}%
\label{C2SR-CPSR-tree}}
\subfloat[DIA-variant]{\>\includegraphics[scale=0.32]{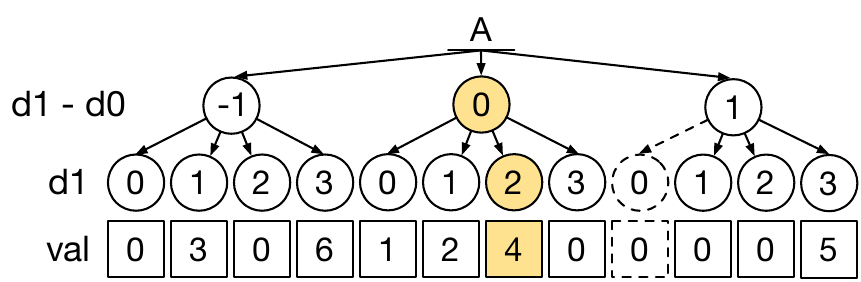}\>
\label{DIA-variant-tree}}
\hfil
\subfloat[DIA]{\>\includegraphics[scale=0.32]{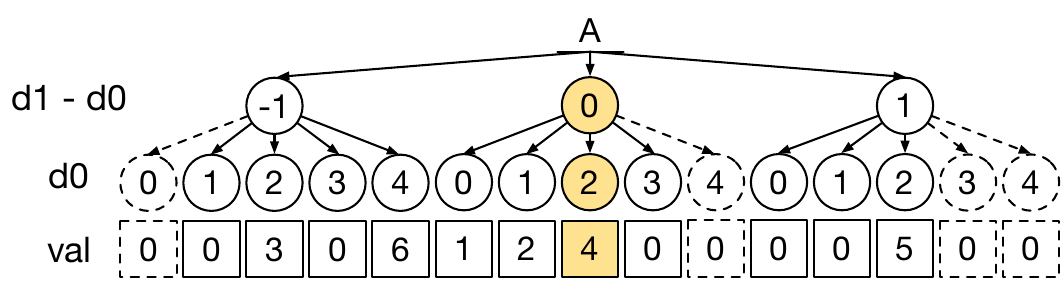}\>
\label{DIA-tree}}
\subfloat[BCSR]{\>\includegraphics[scale=0.32]{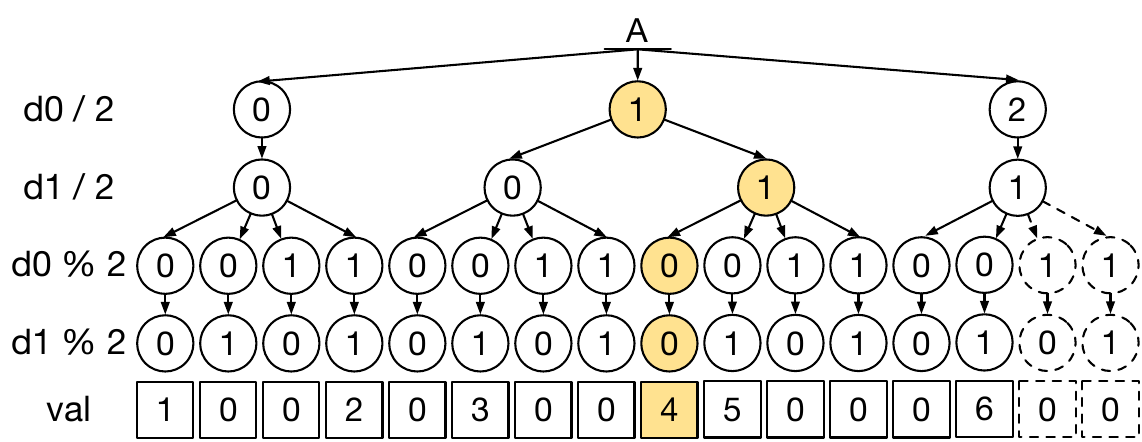}\>
\label{BCSR-tree}}
\hfil
\subfloat[CSB]{\>\includegraphics[scale=0.32]{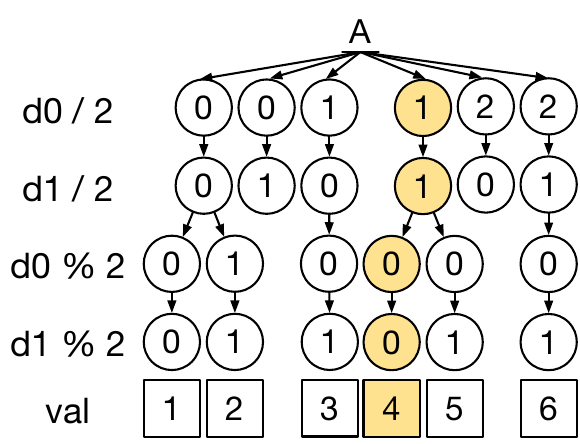}%
\label{CSB-tree}}
\subfloat[ELLPACK]{\>\includegraphics[scale=0.32]{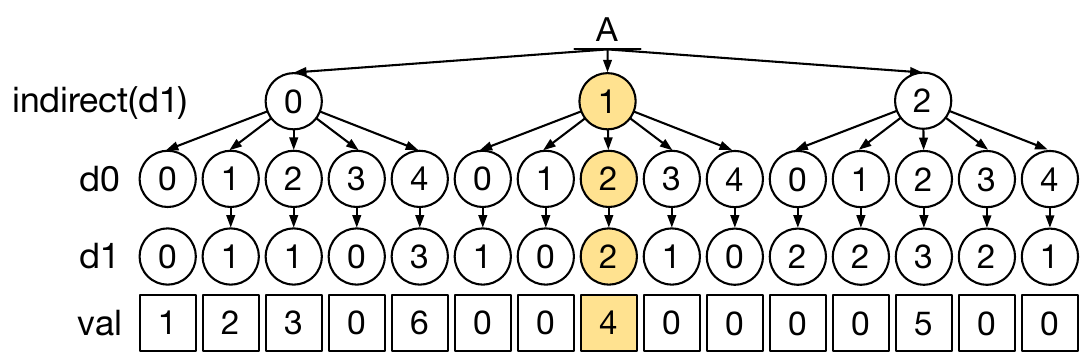}%
\label{ELL-tree}}
\subfloat[CISR]{\>\includegraphics[scale=0.32]{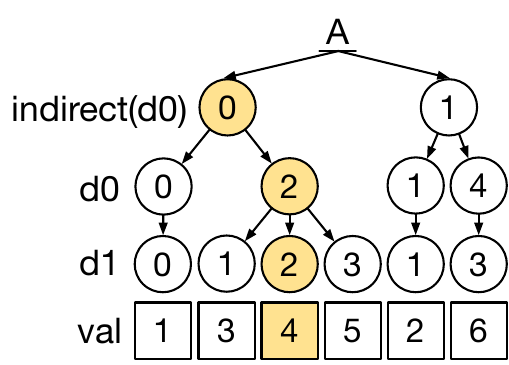}%
\label{CISR-tree}}
\hfil
\caption{The {\Tree}s of the matrix $A$ in Figure \ref{matrix} --- Yellow-shaded blocks refer to the data structure of the same tensor element. The root of a \Tree is the symbol of the tensor. $A_0$ and $A_1$ are sub-matrices of the matrix $A$. The circular nodes are metadata nodes that contain indices, and the arrows represent pointers from a higher major dimension to a lower one. Index expressions are on the left of each coordinate level. Nodes with dashed lines are elements padded for memory alignment.}
\label{fig-index-tree}
\vspace{-10pt}
\end{figure*}

To retain all the information from the original tensor, sparse formats store the metadata of non-zero elements explicitly. The structure of metadata determines the dimension order and intra-dimension coordinate arrangement, which dictate the structure of the corresponding value elements. The metadata structure is represented by a \Tree that stacks tensor dimensions with the major one at the top. Value elements are attached at the bottom of the \Tree, each associated with a path of the tree. In the following, we will introduce how the \Tree is expressed by {\Map}s (\S \ref{coordinate-maps}), query (\S \ref{query}), and mutation (\S \ref{storage-mutation}) primitives. Figure~\ref{fig-index-tree} depicts the {\Tree}s of selected formats presented in Figure \ref{fig:format-storage}.


\subsubsection{{\MAP}}
\label{coordinate-maps}
\rev{We define the index map, denoted by $\mathcal{M}$, as a mapping from a tuple of logical dimension iterators to a tuple of destination index expressions that describe physical dimension iterators:}
\begin{equation*}
    \mathcal{M} \coloneqq \left(d_0, d_1, ..., d_n\right) \mapsto \left(e_0, e_1, ..., e_m\right)
\end{equation*}
\rev{where the logical indices $(d_0, d_1, ..., d_n)$ identify the dimensions of the original tensor, and the physical index expressions $\left(e_0, e_1, ..., e_m\right)$ serve as new dimension identifiers of the sparse format. The \Map of a lossless sparse format has its reverse map 
$\mathcal{M}^{-1} \coloneqq \left(e_0, e_1, ..., e_m\right) \mapsto \left(d_0, d_1, ..., d_n\right)$, which retrieves the dimension iterators of the original tensor.}

{\MAp}s affect the data structures of a sparse format by determining the major order of dimensions and the values of metadata. A simple example of an \Map is \lstinline{(d0, d1) -> (d1, d0)}, which represents a column-major matrix layout. 
In Figure ~\ref{fig-index-tree}, physical index expressions are associated with each level of the \Tree for different formats in Figure~\ref{CSR-tree} - \ref{CISR-tree}, where the original matrix in Figure \ref{matrix} is indexed by \lstinline{d0} and \lstinline{d1}. For example, the CSB format in Figure \ref{CSB-tree} uses an \Map of \lstinline[literate={\\\%}{\%}1]{(d0, d1) -> (d0/2, d1/2, d0\%2, d1\%2)}, where block ids \lstinline{(d0/2, d1/2)} are at the major two levels, and \lstinline[literate={\\\%}{\%}1]{(d0\%2, d1\%2)} are inner block dimensions.
Another example is the DIA format. The traditional DIA format (Figure \ref{DIA-tree}) uses an \Map of \lstinline{(d0, d1) -> (d1-d0, d0)}, which stores elements along the same diagonal with diagonal offsets expressed by \lstinline{(d1-d0)} at the major dimension. 
By changing the second dimension of the \lstinline{dst-index-list} in the traditional DIA format from \lstinline{d0} to \lstinline{d1}, a variant of the DIA format is obtained, as shown in Figure \ref{DIA-variant-tree}. 
In the given example, the DIA-variant format is more space-efficient with \lstinline{d1} ranging from 0 to 3 than the traditional DIA format with \lstinline{d0} ranging from 0 to 4. Compared to TACO \cite{chou2018format}, which only supports the traditional DIA format by hardcoding level functions with types ``range'' and ``offset'', our abstraction is more flexible and expressive by covering a wider range of formats such as DIA-variant via the \Map. 
\begin{figure*}[ht]
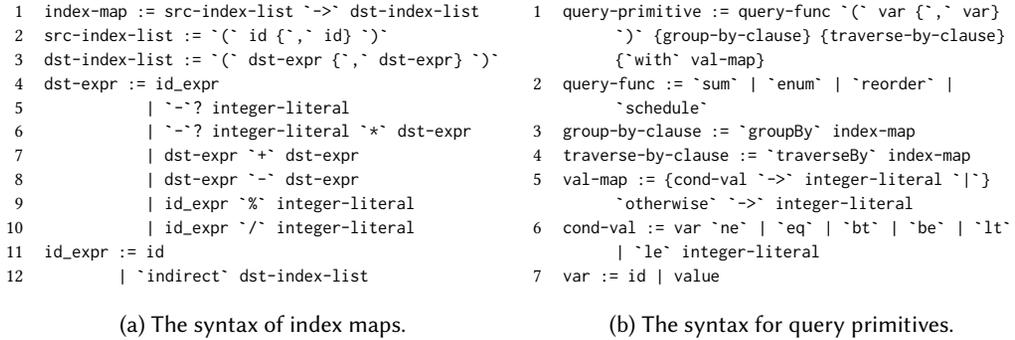

\begin{subfigure}{0.49\textwidth}
\begin{lstlisting}[
language={syntax}
]
index-map := src-index-list `->` dst-index-list
src-index-list := `(` id {`,` id} `)`
dst-index-list := `(` dst-expr {`,` dst-expr} `)`
dst-expr := id_expr  <\label{arith-expr}>
           | `-`? integer-literal
           | `-`? integer-literal `*` dst-expr
           | dst-expr `+` dst-expr
           | dst-expr `-` dst-expr
           | id_expr `%` integer-literal
           | id_expr `/` integer-literal <\label{arith-expr-end}>
id_expr := id
        | `indirect` dst-index-list <\label{indirect-expr}>
\end{lstlisting}
\caption{The syntax of {\Map}s.}
\label{fig-direct-syntax}
\end{subfigure}
\begin{subfigure}{.49\textwidth}
\begin{lstlisting}[
language={syntax}
]
query-primitive := query-func `(` var {`,` var} `)` {group-by-clause} {traverse-by-clause} {`with` val-map}
query-func := `sum` | `enum` | `reorder` | `schedule` <\label{query-func}>
group-by-clause := `groupBy` index-map
traverse-by-clause := `traverseBy` index-map
val-map := {cond-val `->` integer-literal `|`} `otherwise` `->` integer-literal
cond-val := var `ne` | `eq` | `bt` | `be` | `lt` | `le` integer-literal
var := id | value
\end{lstlisting}
\caption{The syntax for query primitives.}
\label{fig-indirect-syntax}
\end{subfigure}
\vspace{-5pt}
\caption{The syntax for {\Map}s and query primitives. \lstinline{id} denotes dimension identifiers, \lstinline{value} denotes tensor element values, and \lstinline{integer-literal} denotes constant numbers. Expressions enclosed in curly braces can be repeated zero or more times.}
\label{fig-syntax}
\vspace{-5pt}
\end{figure*}

\rev{The \IRName syntax of {\Map}s is shown in Figure \ref{fig-direct-syntax}.
The map takes a list of logical dimension indices as inputs and returns a list of arithmetic expressions that represent physical dimension iterators.}
\rev{The physical dimension iterators} are typically expressed as closed-form functions using basic arithmetic operations (e.g., \lstinline{+}, \lstinline{-}, \lstinline{*}, \lstinline{/}, \lstinline[literate={\\\%}{\%}1]{\%}) as shown in Line \ref{arith-expr}-\ref{arith-expr-end} of Figure \ref{fig-direct-syntax}. We refer to {\Map}s with pure arithmetic operations as \emph{direct maps}. 
\rev{For generality, we further allow user-defined custom functions, namely \emph{indirect maps}, for the index iterator, rather than being limited to closed-form expressions. These dimension iterators are marked with the keyword \lstinline{indirect}, followed by their parameter index terms (expressed as \lstinline{dst-index-list}).}
Examples of \emph{indirect maps} can be seen at the major dimensions of several formats, including ELLPACK (Figure \ref{ELL}), CISR (Figure \ref{CISR}), and the BELL portion of the hybrid BELL/COO (Figure \ref{BELL-COO}) format. In the next section, we introduce how indirect map functions are constructed in more detail. 

\subsubsection{Query Primitives}
\label{query}
\IRName provides methods to obtain statistics of a sparse tensor through query primitives. Each query primitive consists of a query function, a \lstinline{group-by} clause, a \lstinline{traverse-by} clause, and a \lstinline{value map}. The syntax of query primitives is shown in Figure \ref{fig-indirect-syntax}. 

The \lstinline{group-by} clause uses an index map with usually fewer dimensions in the destination index expressions to divide tensor elements into groups. Commonly used operators include divide and modulo. For instance, the \lstinline{group-by} function with map \lstinline[literate={\\\%}{\%}1]{(d0, d1) -> (d0\%2)} assigns values in even rows to one group and values in odd rows to another group. Another map \lstinline{(d0, d1) -> (d1-d0)} groups elements on the same diagonals together. The \lstinline{traverse-by} clause specifies the traversing order of the indices with also an index map. For example, \lstinline{traverse-by} \lstinline{(d0, d1) -> (d0)} and \lstinline{(d0, d1) -> (d1)} indicate traversing elements in increasing order of the row and column dimension identifier, respectively. The \lstinline{value map} assigns a new value for each set of values satisfying a certain condition. For example, \lstinline{value ne 0 -> 1 | otherwise -> 0} assigns non-zero elements to 1 and others to 0. 

We predefine several query functions in Line \ref{query-func}: \SumOp, \EnumerateOp, \ReorderOp, and \ScheduleOp. The \SumOp function computes the accumulation of values returned by the \lstinline{value map} with a set of groups defined by the \lstinline{group-by} clause. The \EnumerateOp assigns counting numbers with the start number defined by the \lstinline{value map} in groups that follow the specified traversing order by \lstinline{traverse-by}. The \ReorderOp returns the indices of the sorted elements with a specific traversing order. The \ScheduleOp assigns tensor elements into a specified number of partitions in a balanced manner.

There are primarily two scenarios in \IRName where query primitives are used. The first scenario involves supporting hybrid formats, where the \SumOp function is used to query the \rev{non-zero} distribution patterns before decomposing the input tensor. The other scenario involves supporting custom {\Map}s with indirect mapping functions. Figure \ref{fig-encoding} illustrates how query primitives are used to construct indirect functions. Specifically, \SumOp and \EnumerateOp are used to construct the indirect level of the ELL format in Lines \ref{ELL_indirect}. These functions assign counting numbers to non-zero and zero elements in increasing order. The \SumOp and \ScheduleOp functions constitute the indirect level in the CISR format (Line \ref{CISR_indirect}), which assigns rows to two buckets, each with a balanced number of non-zero elements. 

\subsubsection{Mutation Primitives}
\label{storage-mutation}
While {\Map}s can be used to define new index values and change the dimension order, they do not take advantage of the sparsity to compress the data structures. In this section, we present primitives that enable the compression of tensor data structures. Specifically, we propose two mutation primitives -- \TrimOp, which indicates the removal of tensor components associated with zero-values, and \MergeOp, which expresses the reduction of replicated components.
\rev{We also introduce conversion operators related to the primitives \TrimOp and \MergeOp to better illustrate how the expression of one format can be changed to another format by making slight modifications to its encoding. Note that these conversion operators are distinct from encoding primitives used to describe formats. To differentiate between them, we name conversion operators in CamelCase notation.}

\begin{figure}[t]
    \centering
    \includegraphics[width=0.95\linewidth]{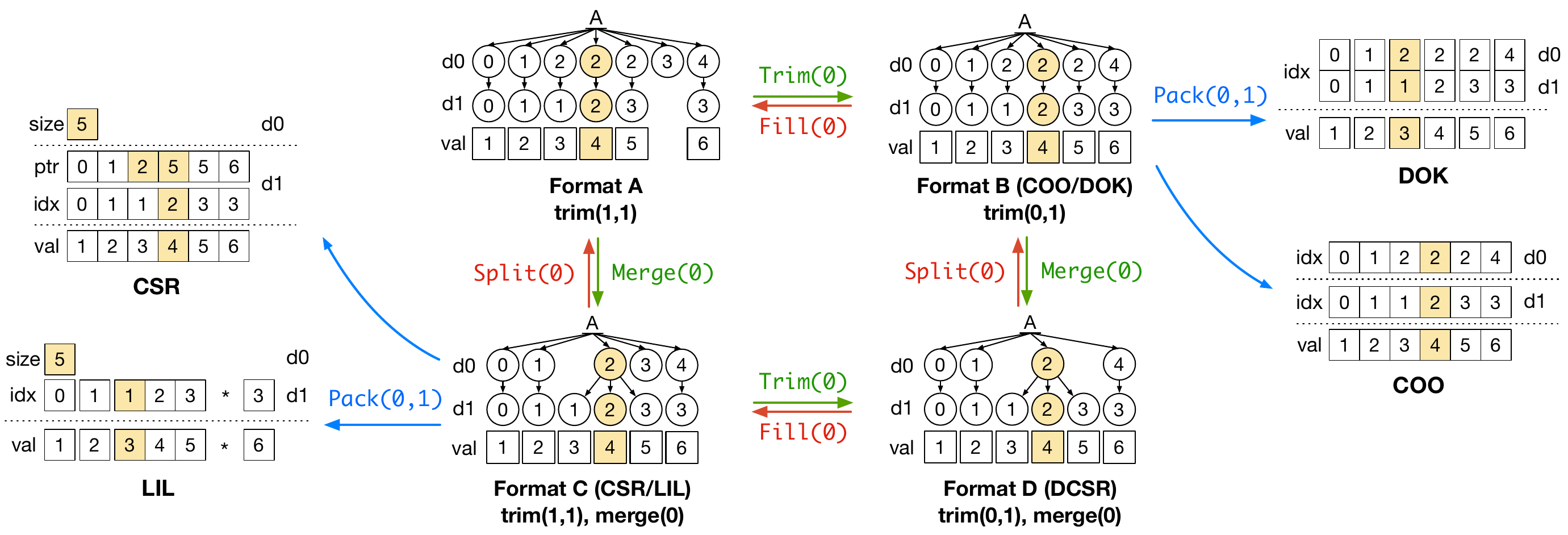}
    \vspace{-5pt}
    \caption{An illustration of mutation primitives and layout primitives. The formats A, B, C, and D of the matrix $A$ in Figure \ref{matrix} can be described using different combinations of the \MergeOp and \TrimOp primitives. Each format can be transformed into its neighbor via a conversion operator indicated by the arrows. Green arrows denote sparsification directions, and red arrows denote densification directions. Notice that the two conversion operators \lstinline{Trim} and \lstinline{Merge} are orthogonal and commutative. Blue arrows point to different physical layouts of the formats.
    } 
    \label{fig:trim-merge}
    \vspace{-10pt}
\end{figure}

\textbf{Trim.} The \TrimOp primitive takes two numerical values representing dimension levels, where the number 0 corresponds to the major dimension at the top of the metadata hierarchy, and subsequent levels increase from top to bottom. The primitive \lstinline{trim(S,E)} indicates the removal of zero values (at the bottom of the tree) and the associated metadata (on the path from the root to leaves) between a starting level \lstinline{S} and an ending level \lstinline{E}, where~\lstinline{S}$\le$\lstinline{E} (i.e., \lstinline{S} is closer to the root than \lstinline{E}). Specifically, \lstinline{trim(S,E)} first removes data nodes belonging to the sub-tensor identified by \lstinline{(d0, d1, ..., dE)}, if all elements in that sub-tensor are zeros. Then it cleans up dangling nodes that have no children nodes from level \lstinline{E} up to level \lstinline{S}. For example, in formats B and D of Figure \ref{fig:trim-merge}, the primitive \lstinline{trim(0,1)} indicates that all zero values identified by dimensions \lstinline{(d0, d1)} are removed, along with all their metadata. In contrast, the primitive \lstinline{trim(1,1)} in formats A and C only removes zero values and their column indices at the least major dimension \lstinline{(d1)}, while leaving dangling nodes (row index 3) at dimension \lstinline{(d0)}.


\rev{The conversion operator \lstinline{Trim(L)} implements the removal of metadata at level \lstinline{L} if the sub-tensor identified by indices at level \lstinline{L} contains only zeros. The reverse operator \lstinline{Fill(L)} inserts nodes at level \lstinline|L| for all existing parents by creating nodes with missing index values in the range of \lstinline|[0,DL)| where \lstinline|DL| is the dimension size at level \lstinline|L|.}
Figure \ref{fig:trim-merge} shows how \lstinline{Fill(0)} converts format B to A and D to C by adding index nodes for the empty row.

\textbf{Merge.} 
The \MergeOp primitive reduces storage space by merging equivalent paths at specified levels. It takes in a list of numerical values representing dimension levels, and for each level \lstinline{L} that has \MergeOp applied, all repeating paths from the top-level dimension down to level \lstinline{L} will be fused into one path. Children nodes at level \lstinline{L+1} sharing the same metadata are inherited by the fused parent node. In Figure \ref{fig:trim-merge}, formats C and D have the encoding \lstinline{merge(0)}, which indicates fusing repetitive nodes at dimension \lstinline{(d0)}.

\rev{The conversion operator \lstinline{Merge(L)} implements the fusion of repetitive metadata nodes at level \lstinline{L}. The reverse operator \lstinline{Split(L)} restores multiple copies of parent paths from the root to level \lstinline{L} for nodes with more than one child at level \lstinline{L}.}
In Figure \ref{fig:trim-merge}, applying \lstinline{Split(0)} converts format C to A and D to B by replicating nodes at the major dimension, ensuring that each row index node has only one child node.

\begin{figure}[t]
\vspace{-2pt}
\begin{minipage}{\columnwidth}
\begin{lstlisting}[
basicstyle={\scriptsize\ttfamily},
identifierstyle={\color{black}},
language={mlir},
tabsize=2,
numbersep=8pt,
numbers=left,
breaklines=true,
xleftmargin=2em, framexleftmargin=2em, aboveskip=1mm, belowskip=1mm,
escapeinside=''
]
#COO = #encoding<{ idx_map<(d0,d1)->(d0,d1)>, mutation<trim(0,1)> }>
#DOK = #encoding<{ idx_map<(d0,d1)->(d0,d1)>, mutation<trim(0,1)>, layout<pack(0,1)> }>
#CSR = #encoding<{ idx_map<(d0,d1)->(d0,d1)>, mutation<merge(0), trim(1,1)> }>
#DCSR = #encoding<{ idx_map<(d0,d1)->(d0,d1)>, mutation<merge(0), trim(0,1)> }>
#DIA = #encoding<{ idx_map<(d0,d1)->(d1-d0,d0)>, mutation<merge(0), trim(0,0)> }>
#DIA-variant = #encoding<{ idx_map<(d0,d1)->(d1-d0,d1)>, mutation<merge(0), trim(0,0)> }>
#BCSR = #encoding<{ idx_map<(d0,d1)->(d0/2,d1/2,d0%2,d1%2)>, mutation<merge(0,1), trim(1,1)> }>
#CSB = #encoding<{ idx_map<(d0,d1)->(d0/2,d1/2,d0%2,d1%2)>, mutation<merge(1), trim(2,3)> }>
#ELL = #encoding<{ idx_map<(d0,d1)->(indirect(d1),d0,d1)>, mutation<merge(0), trim(0,0)>,
  indirect<{ sum(value) groupBy (d0,d1)->(d0) with value ne 0 -> 1 | otherwise -> 0 '\label{ELL_indirect}'
             enum(value) groupBy (d0,d1)->(d0) traverseBy (d0,d1)->(d1) with value eq 0 -> sumVal | otherwise -> 0 }> }>
[#C2SR = #encoding<{ idx_map<(d0,d1)->(d0%2,d0/2,d1)>, mutation<merge(0,1), trim(2,2)>, layout<partition(0)>}>
#CISR = #encoding<{ idx_map<(d0,d1)->(indirect(d0),d0,d1)>, mutation<merge(0,1), trim(1,2)>,
  indirect<{ sum(value) groupBy (d0,d1)->(d0) with value ne 0 -> 1 | otherwise -> 0 '\label{CISR_indirect}'
             schedule(d0) traverseBy (d0,d1)->(d0/2) }>,
  layout<partition(0)> }>
\end{lstlisting}
\end{minipage}
\vspace{-8pt}
\caption{Encodings of selected sparse formats in Figure \ref{fig:format-storage} written in pseudo MLIR code. 
}
\label{fig-encoding}
\vspace{-10pt}
\end{figure}

\subsection{Data Layout}
\label{layout}
A \Tree can be realized using different physical memory layouts. By default, it is stored in an SoA layout, where each dimension is stored using separate arrays. \IRName further provides two primitives to express different memory layout options. The \PartitionOp primitive determines whether to divide one tensor into several sub-tensors, which can be beneficial for storing the sparse tensor in banked memories. The \PackOp primitive switches from the SoA layout to an AoS layout for a single tensor, providing a larger design space with more layout options for users to explore.


\textbf{Partition.} 
The \PartitionOp primitive splits metadata with their descendant nodes at the specified dimension level into subtrees, each stored separately as a sub-tensor. This is particularly useful for multi-bank memory systems, where each sub-tensor can be stored in a different bank to reduce memory access conflicts and improve performance. For example, the C\textsuperscript{2}SR and CISR formats are partitioned into sub-tensors with \lstinline{partition(0)}, and each sub-tensor is designed to occupy a memory bank. The format encodings of the C\textsuperscript{2}SR and CISR formats are shown in Figure \ref{fig-encoding}.

\textbf{Pack.} 
The \PackOp primitive enables the creation of an AoS layout for a tensor by specifying two dimension levels, \lstinline{S} and \lstinline{E}, and enforcing a depth-first traversal of the \Tree for dimensions from \lstinline{S} to \lstinline{E}. By default, a breadth-first traversal of the \Tree generates level-wise arrays which correspond to an SoA format. For example, in Figure \ref{fig:trim-merge}, a breadth-first traversal of the \Tree B results in \lstinline|{d0=[0,1,2,2,2,4], d1=[0,1,1,2,3,3], val=[1,2,3,4,5,6]}|, which corresponds to the COO format. Similarly, the default traversal order of the \Tree C leads to an SoA layout, which is the CSR format. 
On the other hand, adding the primitive \lstinline{pack(0,1)} to the \Tree B pairs up index values at dimensions \lstinline{(d0, d1)} and generates an AoS layout, i.e., \lstinline|[{d0=0, d1=0, val=1}, {d0=1, d1=1, val=2}, ..., {d0=4, d1=3, val=6}]|, corresponding to the DOK format. Similarly, adding the primitive \lstinline{pack(0,1)} to format C generates the AoS layout counterpart of the CSR format, which is the LIL format.
 



With the above primitives, \IRName can express a wide range of custom formats, including those with custom {\Map}s (e.g., load-balanced formats in Figure \ref{CISR}), \rev{sparsity pattern}-aware hybrid formats (e.g., Figure \ref{BDIA-CSR}/\ref{BELL-COO}), and backend-aware memory layouts (e.g., banked formats in Figure \ref{C2SR}/\ref{CISR}). Figure \ref{fig-encoding} lists the abstractions of all the formats in Figure \ref{fig:format-storage}.

\section{Compilation}
\label{codegen}

To generate code for format customization and compute operations, the \IRName compiler first decodes the data structures and layouts from format descriptions specified in the intermediate language (\S \ref{layout-inference}). \rev{Once the source and destination formats are determined, the compiler lowers the \lstinline{convert} operation by applying a sequence of rewrite rules and emitting conversion operators that gradually convert from the source format to the destination format (\S \ref{format-conversion}). For compute operation lowering, \IRName 
leverages the MLIR SparseTensor dialect to handle classic formats with the index map \lstinline{(d0, d1)->(d0, d1)}. In the case of custom formats, \IRName implements a general compute kernel generation algorithm (\S \ref{compute-codegen}). 
}
\vspace{-5pt}

\subsection{Inference of Data Structure and Layout}
\label{layout-inference}
The \IRName compiler infers the data structures and layouts, like those shown in Figure \ref{fig:format-storage}, from the formats specified in the UniSparse language (Figure \ref{fig-encoding}). \rev{The conservative approach for storing a sparse format involves maintaining an array of indices (\lstinline{idx}) and an array of pointers (\lstinline{ptr}) at each level of the \Tree.} An element in an \lstinline{idx} array identifies a node at the current tree level. An element in a \lstinline{ptr} array indicates how many nodes are connected to a parent node, i.e., it encodes a down arrow $\downarrow$ in Figure ~\ref{fig-index-tree}. \rev{However, the storage can be simplified in special cases. For example, when} the indices at the current level are contiguous numbers starting from 0, the \lstinline{idx} array can simply be replaced with a \lstinline{size}. In another case where nodes have a one-to-one correspondence with their parents, the \lstinline{ptr} array can be skipped. 


 
  Based on these principles, the compiler infers the simplified data structures of a sparse format. \rev{A dimension level with no \TrimOp or \MergeOp applied is stored in \lstinline{size}.}
 The \lstinline{trim(S,E)} primitive requires explicit \lstinline{idx} arrays at dimensions from \lstinline{S} to level \lstinline{E}, as indices at these levels are no longer contiguous after being trimmed. The \MergeOp primitive adds a \lstinline{ptr} array to the descendant levels of the specified levels, if the descendant levels also have \TrimOp applied, since the number of children nodes varies from one parent node to another at the level being merged.
 For example, to store the CSR format, an \lstinline{idx} array is required at the column dimension because it is trimmed, and a \lstinline{ptr} array is also needed at the column dimension, as the parent dimension (i.e., the row dimension) is merged. The DCSR format requires an \lstinline{idx} array at both the row and column dimensions since they are both trimmed. The DIA format requires an \lstinline{idx} array at the major dimension, which stores the diagonal offsets. Although the major dimension has been merged, it does not add a \lstinline{ptr} array to its descendant level since the second dimension is not trimmed.
Indirect maps introduce less regular index patterns, which also require an explicit \lstinline{idx} array. 


The aforementioned steps infer the data structures of a sparse format, but the actual physical layout is not yet determined. In \IRName, the physical memory layout of a sparse tensor can be customized using the \PartitionOp and \PackOp primitives. The \PartitionOp primitive divides indices and all their descendant metadata at the specified level into sub-tensors. For example, the C\textsuperscript{2}SR format stores sub-tensors separately in memory banks, with the \PartitionOp primitive applied at the major level. On the other hand, the \lstinline{pack(S,E)} primitive generates an AoS layout of the tensor from level \lstinline{S} to level \lstinline{E}, as opposed to the default SoA layout. The value array at the bottom of the \Tree, which is treated as an additional level of data, can also be packed with metadata. If the packed levels have the same number of elements, they are stored in an array of tuples, such as the DOK format. Alternatively, the elements can be stored in an array of variable-length lists, such as the LIL format.




\vspace{-5pt}

\subsection{Format Conversion}
\label{format-conversion}
 \rev{The \IRName compiler incorporates a general algorithm that automates conversion between any two custom formats specified with mutation primitives and index maps that contain purely arithmetic operations \lstinline[literate={\\\%}{\%}1]|\{+, -, *, \%, /\}|. Indirect functions and layout primitives are not reversible, and therefore, \IRName does not support automated conversion when the {\it source} format contains indirect functions or layout primitives. However, they are allowed in the target format.} 

\rev{The format conversion algorithm within \IRName involves a thorough analysis of both the source and destination format encoding. It proceeds by individually conducting pattern matching on each component of the source and target format encoding.
This process applies a series of rewrite rules to convert the source format step-by-step to the target format. These rewrite rules
are implemented as conversion operators that transform data structures and layouts in atomic steps. 
}

\rev{
Table \ref{table:builtin-primitives} presents a collection of rewrite rules along with their corresponding conversion operators.
These rewrite rules can be categorized into four sets: the rules for rewriting index maps including affine ({\circled[10]{1}}, \circled[10]{2}, and \circled[10]{3}) and non-affine (\circled[10]{4} and \circled[10]{5}) transformations, for matching mutation primitives (\circled[10]{6}-\circled[10]{11}), for querying (\circled[10]{12}-\circled[10]{15}) and for layout transformation (\circled[10]{16}-\circled[10]{17}). 
}

\rev{Affine transformations are equivalently represented as matrices in the table. For instance, consider rule \circled[10]{1}, which matches the case when two dimensions in the source format switch their positions in the target format. The effect of rule \circled[10]{1} is expressed through an elementary transformation matrix as shown in the ``Matrix'' column of the table, and the operator \lstinline{Swap} updates the data structures in the target \Tree. Non-affine transformation rules are applicable when there are \lstinline{`/`} and \lstinline{`\%`} in the source or target index map. These non-affine transformations are implemented using the \lstinline{TileUnion} and \lstinline{TileSplit} operators.}

\rev{Rule \circled[10]{6}-\circled[10]{11} handle the situations when either the source or the target format contains mutation primitives. For example, if dimension $S$ to dimension $E$ are trimmed in the source format, but the preceding dimension $S-1$ is additionally trimmed in the target format, the corresponding conversion operator should perform this additional trim (Rule \circled[10]{6}). Conversely, if the target format has one less dimension trimmed than the source format, the corresponding conversion operator should fill that dimension (Rule \circled[10]{7}). Similarly, the formats could differ in the last trimmed dimension, and these cases are handled by Rule \circled[10]{8} and \circled[10]{9}. These two rules 
also cover the special situation when $S == E$, in which case one format contains \lstinline{trim} and the other does not. Furthermore, Rule \circled[10]{10} merges one more dimension based on the source format, and Rule  \circled[10]{11} removes one \newrev{merged} dimension.}

\rev{Finally, Rule \circled[10]{12}-\circled[10]{17} directly work on the target format when the target format contains query or layout primitives. Note that these two sets of rewrite rules, as well as the encoding primitives, are open-ended and can be extended when necessary to support new user-defined custom formats.} 

\newrev{
The rewrite rules presented in Table \ref{table:builtin-primitives} ensure that \IRName is fully capable of converting formats encoded only with direct index maps and mutation primitives (referred to as set ${S}$) into arbitrary formats expressed in \IRName notation (referred to as set ${G}$). To demonstrate the completeness of format conversion in \IRName, we first establish its ability to convert between any two formats within set ${S}$ using the rewrite rules provided in Table \ref{table:builtin-primitives}. Then, we discuss the conversion from a format in set ${S}$ to a format in set ${G-S}$ that involves indirect functions and layout primitives.}

\newrev{
We discuss the conversion between arbitrary formats in set $S$ in two steps. }

\newrev{
(i) \textit{Reversibility}. The conversion process between formats within set $S$ is {reversible}, implying that for any two formats, A and B, encoded solely using direct index maps and mutation primitives, if there exists a sequence of conversion operators transforming format A to format B, there must also exist a conversion path that can transform format B back to format A. This reversibility property is guaranteed by the conversion operators outlined in Table \ref{table:builtin-primitives}. 
Specifically, Table \ref{table:builtin-primitives} contains arithmetic conversion operators, each having its reverse pair: \lstinline|Swap(i, j)| and \lstinline|Swap(j, i)|, \lstinline|Scale(i, f)| and \lstinline|Scale(i, 1/f)|, \lstinline|Skew(i, j, f)| and \lstinline|Skew(i, j, -f)|, \lstinline|TileUnion(i, f)| and \lstinline|TileSplit(i, f)|. Furthermore, the conversion operators for rule \circled[10]{6} and \circled[10]{7}, those for rule \circled[10]{8} and \circled[10]{9}, and rule \circled[10]{10} and rule \circled[10]{11} also form reverse pairs.}

\newrev{
(ii) \textit{Reachability}. 
We further show that there exists a reference format, denoted as $R$, from which all formats within set $S$ can be reached, meaning they can be converted from $R$.
Without loss of generality, we select the COO format as $R$. Given the reversibility property of format conversion within set $S$, we only need to show that any arbitrary format within set $S$ can be converted into the COO format. 
This can be achieved through the following observations: 
Rule \circled[10]{5} eliminates pairs of divide and modulo operations, ensuring that any formats within set $S$ can be converted into those without divide and modulo operations in the index maps; 
rules \circled[10]{1} through \circled[10]{3} cover the entire affine transformation space, allowing for the conversion of affine index expressions into the original dimension identifiers in the COO format; and rule \circled[10]{11} removes merged levels, while rules \circled[10]{6} and \circled[10]{8} expand trimmed levels, enabling the conversion of any mutation primitives into the encoding form of COO format.}

\newrev{
Combining the two key points (i) and (ii), we can conclude that \IRName can convert between any format within set $S$ using the rewrite rules and conversion operators in Table \ref{table:builtin-primitives}.
}

\newrev{
For the conversion from a format within set $S$ to a format within set $G-S$, which includes indirect functions and layout primitives, one can directly apply rules \circled[10]{12} through \circled[10]{17} that align a source format with the target format's encoding. Building upon the earlier discussion on format conversion within set $S$, 
we know that if there exists one format within set $S$ capable of converting to any format within $G-S$, then any arbitrary format from set $S$ will have the ability to convert to any format in $G-S$ as well. As a result, \IRName can perform conversions from any format in set $S$ to any format in set $G$.
}

\begin{table}[t]
\centering
\caption{\rev{Rewrite rules and conversion operators that support the \IRName format conversion algorithm. Parameters of query primitives are omitted as \lstinline|\{*\}| due to the limited length of the table. The syntax of the query primitives is shown in Figure \ref{fig-indirect-syntax}, and examples can be found in Figure \ref{fig-encoding}.}}
\vspace{-5pt}
\footnotesize
\setlength\tabcolsep{1pt} 
\lstset{%
tabsize=2,
columns=fullflexible,
basicstyle=\ttfamily\scriptsize,
breaklines=true,
breakatwhitespace,
escapeinside={(*}{*)},
alsoletter={\., \_},
morekeywords={tile\_split, tile\_union, enumerate, reorder, schedule, move, trim, fill, merge, split, vectorize, devectorize, pad, pack, partition, sort, skew, swap, scale}
}
\resizebox{0.85\textwidth}{!}{
\begin{tabular}{c|c|c|c|c|c}
\toprule
\multirow{2}{*}{\textbf{Type}} & \multirow{2}{*}{\textbf{Rule}} & \multirow{2}{*}{\textbf{Source}} & \multirow{2}{*}{\textbf{Target}} & \multicolumn{2}{c}{\textbf{Conversion}} \\\cline{5-6}
 & & & & Matrix & Operator \\\midrule
 \multirow{8}{*}{\rotatebox[origin=c]{90}{\textbf{Index Map}}}
 &
 \circled[10]{1}
 &
$\mbox{\tiny$\begin{psmallmatrix}
d_0, & ..., & d_i, & ..., & d_j, & ..., & d_N
\end{psmallmatrix}$}$
&
$\mbox{\tiny$
\begin{psmallmatrix}
d_0, & ..., & d_j, & ..., & d_i, & ..., & d_N
\end{psmallmatrix}$}$
&
$\mbox{\tiny$\begin{psmallmatrix}
1 &        &        &        &        & & \\
  & \ddots &        &        &        & & \\
  &        & 0 & \dots & 1 &  &  \\
  &        & \vdots & \ddots & \vdots & & \\
  &        & 1      & \dots  & 0      & & \\
  &        &        &        &        & \ddots & \\
  &        &        &        &        &        & 1 
\end{psmallmatrix}$}$
&
\lstinline{Swap(i,j)}
\\\cline{2-6}
&
\circled[10]{2}
&
$\mbox{\tiny$\begin{psmallmatrix}
d_0, & ..., & d_i, & ..., & d_N
\end{psmallmatrix}$}$
&
$\mbox{\tiny$
\begin{psmallmatrix}
d_0, & ..., & f*d_i, & ..., & d_N
\end{psmallmatrix}$}$
&
$\mbox{\tiny$\begin{psmallmatrix}
1 &        &        &        &        \\
  & \ddots &        &        &        \\
  &        & f      &        &   \\
  &        &   & \ddots &   \\
  &        &      &    & 1      
\end{psmallmatrix}$}$
&
\lstinline{Scale(i,f)}
\\\cline{2-6}
&
\circled[10]{3}
&
$\mbox{\tiny$\begin{psmallmatrix}
d_0, & ..., & d_i, & ..., & d_j, & ..., & d_N
\end{psmallmatrix}$}$
&
$\mbox{\tiny$
\begin{psmallmatrix}
d_0, & ..., & d_i, & ..., & f*d_i + d_j, & ..., & d_N
\end{psmallmatrix}$}$
&
$\mbox{\tiny$\begin{psmallmatrix}
1 &        &        &        &        & & \\
  & \ddots &        &        &        & & \\
  &        & 1 & \dots & 0 &  &  \\
  &        & \vdots & \ddots & \vdots & & \\
  &        & f      & \dots  & 1      & & \\
  &        &        &        &        & \ddots & \\
  &        &        &        &        &        & 1 
\end{psmallmatrix}$}$
&
\lstinline{Skew(i,j,f)}
\\\cline{2-6}
&
\circled[10]{4}
&
$\mbox{\tiny$\begin{psmallmatrix}
d_0, & ..., & d_i~ /~ f, & d_i~ \%~ f, & ..., & d_N
\end{psmallmatrix}$}$
&
$\mbox{\tiny$
\begin{psmallmatrix}
d_0, & ..., & d_i, & ..., & d_N
\end{psmallmatrix}$}$
&
\multicolumn{2}{c}{\lstinline{TileUnion(i,f)}}
\\\cline{2-6}
&
\circled[10]{5}
&
$\mbox{\tiny$\begin{psmallmatrix}
d_0, & ..., & d_i, & ..., & d_N
\end{psmallmatrix}$}$
&
$\mbox{\tiny$
\begin{psmallmatrix}
d_0, & ..., & d_i~ /~ f, & d_i~ \%~ f, & ..., & d_N
\end{psmallmatrix}$}$
&
\multicolumn{2}{c}{\lstinline{TileSplit(i,f)}}\\
\bottomrule
\end{tabular}
}
\resizebox{0.85\textwidth}{!}{
\begin{tabular}{c|c|l|l|c?c|c|l|c}
{\textbf{Type}} & {\textbf{Rule}} & {\thead{\textbf{Source}}} & {\thead{\textbf{Target}}} & {\textbf{Conversion}} & {\textbf{Type}} & {\textbf{Rule}} & {\thead{\textbf{Target}}} & {\textbf{Conversion}} \\\cmidrule{1-5}\cmidrule{6-9}
\multirow{8}{*}{\rotatebox[origin=c]{90}{\textbf{Mutation}}}
&
\circled[10]{6}
&
 $\mbox{\tiny$trim(d_{S}, d_{E}), S\leq E$}$
&
 $\mbox{\tiny$trim(d_{S-1}, d_{E}), S\leq E$}$
&
\lstinline{Trim(S-1)}
& 
\multirow{4}{*}{\rotatebox[origin=c]{90}{\textbf{Query}}}
& 
\circled[10]{12}
& $\mbox{\tiny$sum\{*\}$}$ & \lstinline{Sum()} 
\\\cline{2-5}\cline{7-9}
&
\circled[10]{7}
&
 $\mbox{\tiny$trim(d_{S-1}, d_{E}), S\leq E$}$
&
$\mbox{\tiny$trim(d_{S}, d_{E}), S\leq E$}$
&
\lstinline{Fill(S-1)}
& 
&
\circled[10]{13}
& $\mbox{\tiny$enum\{*\}$}$ & \lstinline{Enumerate()} 
\\\cline{2-5}\cline{7-9}
&
\multirow{2}{*}{\circled[10]{8}}
&
\multirow{2}{*}{$\mbox{\tiny$trim(d_S, d_{E-1}), S\leq E$}$}
&
\multirow{2}{*}{$\mbox{\tiny$trim(d_S, d_{E}), S\leq E$}$}
&
\lstinline{Devectorize(E),}
& 
&
\circled[10]{14}
& $\mbox{\tiny$reorder\{*\}$}$ & \lstinline{Reorder()} 
\\\cline{7-9}
&
&
&
&
\lstinline{Trim(E), Vectorize(E+1)}
& 
&
\circled[10]{15}
& $\mbox{\tiny$schedule\{*\}$}$ & \lstinline{Schedule()} 
\\\cline{2-5}\cline{6-9}
&
\multirow{2}{*}{\circled[10]{9}}
&
\multirow{2}{*}{$\mbox{\tiny$trim(d_S, d_{E}), S\leq E$}$}
&
\multirow{2}{*}{$\mbox{\tiny$trim(d_S, d_{E-1}), S\leq E$}$}
&
\lstinline{Devectorize(E+1),}
& 
\multirow{4}{*}{\rotatebox[origin=c]{90}{\textbf{Layout}}}
&
\multirow{2}{*}{\circled[10]{16}}
& \multirow{2}{*}{$\mbox{\tiny$pack(d_S, d_E)$}$} & \multirow{2}{*}{\lstinline{Pack(S, E)}}
\\
&
&
&
&
\lstinline{Fill(E), Vectorize(E)}
& 
& 
& 
& 
\\\cline{2-5}\cline{7-9}
&
\circled[10]{10}
&
{$\mbox{\tiny$merge(d_0, ..., d_{i-1})$}$}
&
{$\mbox{\tiny$merge(d_0, ..., d_{i-1}, d_i)$}$}
&
\lstinline{Merge(i)}
& 
&
\multirow{2}{*}{\circled[10]{17}}
& \multirow{2}{*}{$\mbox{\tiny$partition(d_i)$}$} & \multirow{2}{*}{\lstinline{Partition(i)}} 
\\\cline{2-5}
&
\circled[10]{11}
&
{$\mbox{\tiny$merge(d_0, ..., d_{i-1}, d_i)$}$}
&
{$\mbox{\tiny$merge(d_0, ..., d_{i-1})$}$}
&
\lstinline{Split(i)}
& 
& 
& & 
\\
\bottomrule
\end{tabular}
}
\label{table:builtin-primitives}
\end{table}
\begin{figure}[t]
\begin{minipage}{\columnwidth}
\begin{lstlisting}[language=emit, 
columns=flexible, 
frame=tb,
numbers=left,
basicstyle=\ttfamily\scriptsize,
escapeinside={(*}{*)}]
Algorithm: Format Conversion
Input: source format encoding S, target format encoding T, source format storage S_format
Output: target format storage T_format
FormatConversion {
  T_format = S_format // directly manipulate the source format storage 
  if index_map of S != index_map of T: // step 1: Index Map Alignment (*\label{cvt-line-6}*)
    if index_map of S contains (di/f, di%f): (*\label{cvt-line-7}*)
      apply Rule (*\circled[10]{4}*) to T_format;  (*\label{cvt-line-8}*)
    transMatrix = indexMapMatrix(S)(*$^{-1}$*) * indexMapMatrix(T); (*\label{cvt-line-9}*)
    for each non-identity sub-matrix E of transMatrix: (*\label{cvt-line-10}*)
      apply E to T_format; // Rule (*\circled[10]{1}*),(*\circled[10]{2}*),(*\circled[10]{3}*) in Table (*\ref{table:builtin-primitives}*) (*\label{cvt-line-11}*)
    if index_map of T contains (di/f, di%f): (*\label{cvt-line-12}*)
      apply Rule (*\circled[10]{5}*) to T_format; (*\label{cvt-line-13}*)
    for each indirect_function f of {sum, enum, reorder, schedule} in T: (*\label{cvt-line-14}*)
      apply f to T_format; // Rule (*\circled[10]{12}*)-(*\circled[10]{15}*) in Table (*\ref{table:builtin-primitives}*) (*\label{cvt-line-15}*)
    apply Sort to T_format; // sort indices (*\label{cvt-line-16}*)
  while trim of S != trim of T: // step 2: Structure Mutation (*\label{cvt-line-17}*)
    apply Rule (*\circled[10]{6}*)/(*\circled[10]{7}*)/(*\circled[10]{8}*)/(*\circled[10]{9}*) to T_format;  (*\label{cvt-line-18}*)
  while merge of S != merge of T: (*\label{cvt-line-19}*)
    apply Rule (*\circled[10]{10}*)/(*\circled[10]{11}*) to T_format; (*\label{cvt-line-20}*) 
  for each layout l of {pack, partition} in T: // step 3: Layout Generation (*\label{cvt-line-21}*)
    apply l to T_format; (*\label{cvt-line-22}*) // Rule (*\circled[10]{16}*),(*\circled[10]{17}*) in Table (*\ref{table:builtin-primitives}*)
  return T_format;
}
\end{lstlisting}
\end{minipage}
\vspace{-8pt}
\caption{\rev{The pseudo-code of the \IRName format conversion algorithm.
}
}
\vspace{-10pt}
\label{alg-conversion}
\end{figure}

\rev{Figure \ref{alg-conversion} shows the pseudocode of our format conversion algorithm, which includes three steps. Initially, the compiler aligns the source index map with the target index map (Step 1), followed by the mutation of data structures (Step 2). Finally, the memory layout is generated (Step 3).}

\rev{
In the first step, new indices are calculated for the target format (line \ref{cvt-line-6} - \ref{cvt-line-16}). If the source format encoding \lstinline[morekeywords=S]{S} contains \lstinline{`/`} and \lstinline{`\%`} operations, the compiler employs \lstinline{TileUnion} to eliminate these non-affine operations and generate an affine index map of \lstinline[morekeywords=S]{S} (line \ref{cvt-line-7} - \ref{cvt-line-8}). 
Next, the compiler computes the transformation matrix \lstinline{transMatrix} by multiplying the inverse matrix of the \newrev{source} affine index map with the matrix of the \newrev{target} affine index map (line \ref{cvt-line-9}). The function \lstinline{indexMapMatrix} outputs the matrix representation of an affine index map, e.g., \lstinline{(1, 1; -1, 1)} for \lstinline{(d0+d1, d1-d0)}. By breaking down the transformation matrix \newrev{into elementary transformation matrices} (line \ref{cvt-line-10}), \newrev{the corresponding} conversion operators are sequentially applied to the source format (line \ref{cvt-line-11}). Following this, the compiler applies \lstinline{TileSplit} if the target index map contains tiling (line \ref{cvt-line-12} - \ref{cvt-line-13}). 
The handling of indirect functions (line \ref{cvt-line-14} - \ref{cvt-line-15}) adheres to the encoding sequence of the target format. Now all the target indices are ready, and a \lstinline{Sort} step orders the indices as specified in the target format. In the second step, the compiler addresses the disparities between \lstinline{trim} and \lstinline{merge} primitives in the source and target formats, thereby mutating data structures (line \ref{cvt-line-17} - \ref{cvt-line-20}) using the conversion rules \circled[10]{6} - \circled[10]{11} outlined in Table \ref{table:builtin-primitives}. In the final step, the compiler handles the \lstinline{pack} and \lstinline{partition} in the target format and generates custom layouts (line \ref{cvt-line-21} - \ref{cvt-line-22}). }

\rev{
Examples of conversions from COO to BDIA and from COO to CSR are shown in Figure \ref{fig:illustrate-emitted-operators}. When converting from COO to BDIA format, the process begins with the application of index map rules to calculate a new set of index iterators in the target format: \lstinline{Skew(0, 1, -1)} transforms the COO index map \lstinline{(d0, d1)} into \lstinline{(d0, d1-d0)}, and \lstinline{TileSplit(0, 3)} subsequently divides \lstinline{d0} into tiling dimensions \lstinline{d0/3} and \lstinline{d0\%3}. To ensure the proper metadata order \lstinline{(d0/3, d1-d0, d0\%3)}, the \lstinline{Sort} operator is employed. Following index map alignment, the two-dimensional format COO turns into a three-dimensional intermediate format with a mutation primitive \lstinline{trim(0,1)}. To convert the intermediate format into the target BDIA format that would contain  \lstinline|\{merge(0), trim(1,1)\}|, \lstinline{Fill(0)} is applied to obtain \lstinline|trim(1,1)|, and \lstinline{Merge(0)} is used to derive \lstinline|merge(0)|. In the case of COO to CSR format conversion, the distinction lies only in their mutation primitives. Therefore, converting from COO to CSR involves just two operators: \lstinline{Fill(0)}, which matches the pattern converting from \lstinline{trim(0,1)} in COO to \lstinline{trim(1,1)} in CSR, and \lstinline{Merge(0)}, which transforms the absence of merge in COO to \lstinline{merge(0)} in CSR.
}

\subsection{Compute Kernel Generation}
\label{compute-codegen}
\begin{figure*}[ht]
\begin{subfigure}{0.36\columnwidth}
    \centering
    \includegraphics[scale=0.295]{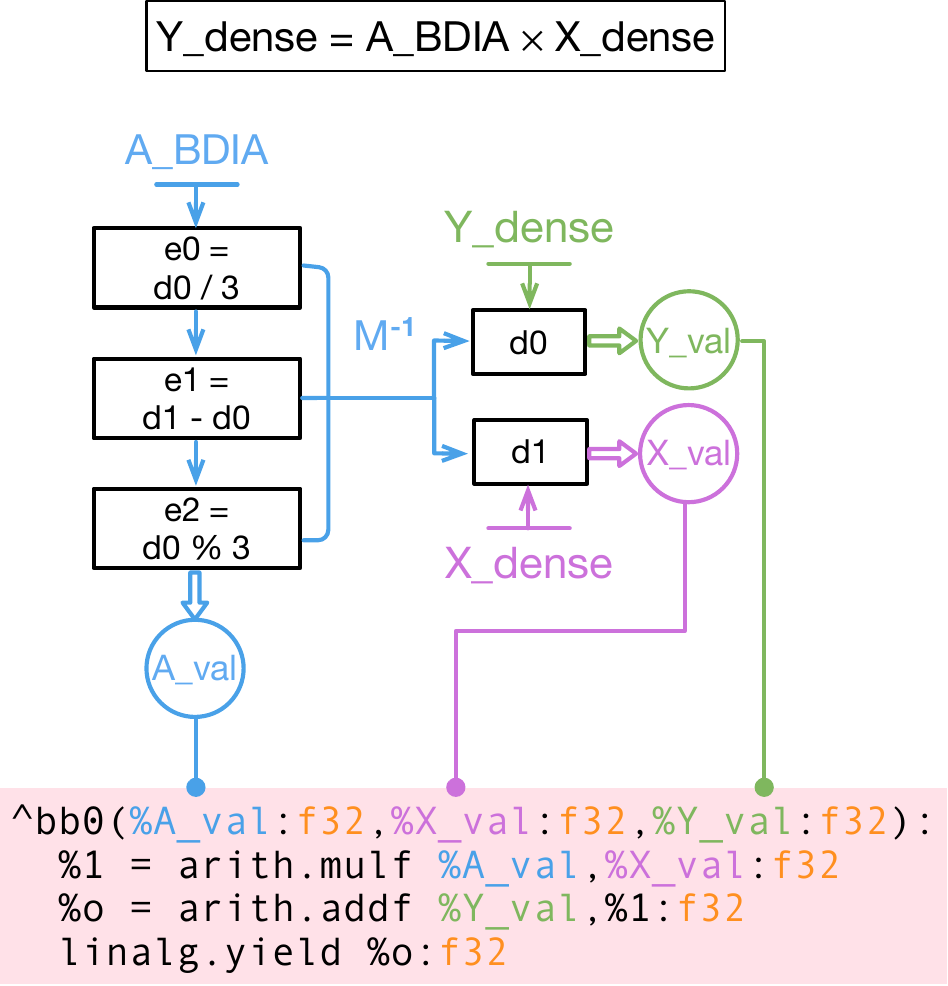}
    \caption{Logical index matching among \\tensor operands.}
    \label{fig:BDIA-codegen}
\label{codegen-example-a}
\end{subfigure}
\begin{subfigure}{0.63\columnwidth}
\begin{lstlisting}[language=codegen, 
columns=flexible, 
frame=tb,
tabsize=2,
numbersep=8pt,
numbers=left,
breaklines=true,frame=tb,
xleftmargin=2em, framexleftmargin=2em, aboveskip=1mm, belowskip=1mm,
basicstyle={\scriptsize\ttfamily},
escapeinside={(*}{*)}]
for id0 in range(0,bdia_size_0): (*\label{e-line-1}*)
  for id_1_0 in range(0,bdia_ptr_1.len()-1): (*\label{e-line-2}*)
    for id_1_1 in range(bdia_ptr_1[id_1_0],bdia_ptr_1[id_1_0+1]): (*\label{e-line-3}*)
      for id2 in range(0,bdia_size_2):  (*\label{e-line-4}*)
        for x_id in range(0,X_dense.len()): // step 2: remove  (*\label{e-line-5}*)
          for y_id in range(0,Y_dense.len()): // step 2: remove(*\label{e-line-6}*)
            bdia_e0 = id0 (*\label{e-line-7}*)
            bdia_e1 = bdia_idx_1[id_1_1] (*\label{e-line-8}*)
            bdia_e2 = id2 (*\label{e-line-9}*)
            bdia_d0 = bdia_e0*3 + bdia_e2(*\label{e-line-10}*)
            bdia_d1 = bdia_e1 + bdia_d0 (*\label{e-line-11}*)
            X_dense_d1 = x_id // step 2: X_dense_d1 = bdia_d1
            Y_dense_d0 = y_id // step 2: Y_dense_d0 = bdia_d0 (*\label{e-line-13}*)
            if (bdia_d0 in range(0,D0) and bdia_d1 in range(0,D1)  (*\label{e-line-14}*)
                and bdia_d1 == X_dense_d1): // step 2: remove (*\label{e-line-15}*)
              A_val = bdia_val[id_1_1 * bdia_size_2 + id2](*\label{e-line-16}*)
              X_val = X_dense[X_dense_d1]
              Y_val = Y_dense[Y_dense_d0] (*\label{e-line-18}*)
              res = Y_val + A_val*X_val (*\label{e-line-19}*)
              Y_dense[Y_dense_d0] = res (*\label{e-line-20}*)
\end{lstlisting}
\caption{The pseudo-code of the generated SpMV kernel.}
\label{codegen-example-b}
\end{subfigure}
\caption{Exemplifying the \IRName kernel generation algorithm with the SpMV kernel using BDIA format. 
}
\vspace{-8pt}
\label{codegen-example}
\end{figure*}

\rev{\IRName extends the MLIR SparseTensor dialect, which leverages TACO's code generation algorithms, to generate compute kernels for conventional formats with the index map \lstinline{(d0, d1) -> (d0, d1)}. When dealing with formats with custom index maps that are unsupported by MLIR SparseTensor, \IRName adopts a two-step code generation algorithm, where the compiler generates a functionally
correct code in the first step, and optimizes the generated code for better performance in the second step. Figure \ref{codegen-example} illustrates the \IRName kernel generation algorithm using the SpMV kernel in BDIA format as an example.}

\rev{In the first step, the \IRName compiler generates a compute kernel that conducts exhaustive iterations across all dimensions of all operands (i.e., tensors of both inputs and output) and performs computations on values with matching indices.
The rationale behind generating an exhaustive iteration across all metadata dimensions is rooted in the fact that different physical dimension iterators of tensor operands can not be co-iterated. For instance, consider the SpMV case with the BDIA format, as illustrated in Figure \ref{fig:BDIA-codegen}. Here, the matrix A features dimension iterators expressed as \lstinline{(e0=d0/3, e1=d1-d0, e2=d0\%3)}, while the input vector has iterator \lstinline{(d0)} and the output vector has iterator \lstinline{(d1)}. There exist five distinct physical dimension iterators \lstinline|\{e0, e1, e2, d0, d1\}|, but no pairs can be iterated together. Therefore, as illustrated in Figure \ref{codegen-example-b}, the \IRName compiler generates a loop nest that iterates through all the dimensions of the input tensor (line \ref{e-line-1} - \ref{e-line-4}), the input vector (line \ref{e-line-5}), as well as the output vector (line \ref{e-line-6}). The iteration through the dimensions of the tensor corresponds to traversing the metadata tree. Progressing from one tree level (i.e., a tensor dimension) to the next level requires traversing a single tree edge, which may involve memory accesses.
In this example, there are 3 tree levels (see the BDIA example in Figure~\ref{BDIA-CSR}): the first and last levels are encoded by two sizes, each visited by a single loop (lines \ref{e-line-1} and \ref{e-line-4}, respectively), and the indices of the tensor for these two levels are simply the loop variables (line \ref{e-line-7} and \ref{e-line-9}); however, the middle level is encoded by an index array and a pointer array, requiring it to be visited in two loops (line \ref{e-line-2}-\ref{e-line-3}), and the tensor's index for this level is retrieved from the pointer array via a memory access (line \ref{e-line-8}).
}

\rev{Now that the indices of the tensor in BDIA format have been retrieved, the logical indices can be restored (lines \ref{e-line-10} - \ref{e-line-11}). In general, the \IRName compiler employs the reverse transformation $\mathcal{M}^{-1}$ to restore the logical indices, as illustrated in Figure \ref{fig:BDIA-codegen}.
In this specific example, $\mathcal{M}^{-1}$ comprises the transformations \lstinline[literate={\\\%}{\%}1]{d0=(d0/3)*3+(d0\%3)} and \lstinline|d1=(d1-d0)+d0|. Once the logical indices are recovered, the compiler generates code to verify whether these logical indices fall within valid boundaries (line \ref{e-line-14}), and ensures the index values of the contraction dimensions match (line \ref{e-line-15}). In this case, the contraction dimension is \lstinline{d1}, so the generated code checks for the equality of index \lstinline{d1} shared by the matrix $A$ and the input vector $X$. 
Inside the loop body, tensor element values are accessed (lines \ref{e-line-16} - \ref{e-line-18}), followed by computation (line \ref{e-line-19}) which is lowered from the program specification in Figure \ref{fig:BDIA-codegen}, and the results are written into the output tensor (line \ref{e-line-20}).}

\rev{In the second step, the \IRName compiler removes redundant iterations from the generated kernel for better performance. To merge iterations, the \IRName compiler checks for several cases:} 
\begin{enumerate}[leftmargin=*,topsep=0pt,label=\rev{\textbf{\alph*.}}]
\item \rev{Co-iteration within a single tensor -- consecutively trimmed dimensions contain metadata of the same length, enabling them to be co-iterated. For instance, in the COO format with \lstinline{trim(0,1)}, both the row and column dimensions contain index arrays of identical lengths, allowing them to be iterated together using a single loop.}
\item \rev{Co-iteration across multiple tensors -- \\
\textbf{Case I.} The contraction dimension of multiple tensors can be co-iterated if they share the same physical dimension iterator. For example, when multiplying DIA (index map \lstinline|(d0, d1) -> (d1-d0, d0)|) with DCSR (index map \lstinline|(d0, d1) -> (d0, d1)|), the second dimension of DIA and the row dimension of DCSR share the physical iterator \lstinline{d0}, allowing these two dimensions to be co-iterated as $~$\lstinline[language=codegen]{while (dia_id1 < dia_id1.len() and dcsr_id0 < dcsr_idx_0.len())}.}\\
\rev{\textbf{Case II.} Dimensions that support random access can reuse the shared iterator of other tensors. For example, in Figure \ref{codegen-example}, the input and output vectors are dense arrays that can be randomly accessed, enabling them to directly borrow the logical index iterators of the matrix A. Consequently, the compiler eliminates iterations in lines \ref{e-line-5}-\ref{e-line-6} and the index equality check in line \ref{e-line-15}. } 
\end{enumerate}
\vspace{-10pt}

\section{The \IRName MLIR Dialect}
\label{IR}
The \IRName intermediate language is implemented as a standalone MLIR dialect. Figure \ref{fig:illustrate-unisparse-code} shows a \IRName program of the SpMV kernel with the hybrid BDIA/CSR format. This section introduces the key ingredients of the \IRName language:

\textbf{Types and Attributes.} Format descriptions are specified as MLIR attributes. In the example program shown in Figure \ref{fig:illustrate-unisparse-code}, all formats are specified beforehand using {\Map}s (idx\_map) and mutation primitives (mutation)
in Line~\ref{spmv-format-abstraction}-\ref{spmv-format-abstraction-end}. 
Sparse tensors are declared with \lstinline{tensor} types and sparse format encoding attributes. Annotations specify the desired sparse format, eliminating the need for explicit handling of sparsity in code. 

\textbf{Tensor Preprocessing Operations.}
The \IRName MLIR dialect defines two key operations for sparse format customization: \DecomposeOp and \ConvertOp. 
The \DecomposeOp operation splits the input tensor into sub-tensors with different sparsity ranges, based on the \rev{non-zero} distribution patterns (identified by the \SumOp primitive).\rev{The \ConvertOp operation specifies format conversion between source and destination formats based on their respective encodings. }

\textbf{Compute Kernels.} The compute kernel is specified using the \lstinline{linalg.generic} operation within MLIR. \rev{Figure \ref{fig:illustrate-unisparse-code} (lines \ref{spmv-signature}-\ref{spmv-compute-end}) provides an example of the \lstinline{linalg.generic} operation that specifies an SpMV kernel. Within this operation, the \#spmv attribute (lines \ref{spmv-signature}-\ref{spmv-signature-end}) indicates the logical dimension iterators and compute patterns of tensor operands. The inputs and outputs (lines \ref{spmv-linalg-ins}-\ref{spmv-linalg-generic}) of the \lstinline{linalg.generic} operation define the iteration space, while the loop body (lines \ref{spmv-compute-start}-\ref{spmv-compute-end}) details the computational logic.}


\section{Evaluation}
\label{evaluation}
This section demonstrates the efficacy of \IRName through a series of case studies (\S \ref{eval-customization}). These case studies illustrate how the adoption of custom formats enabled by \IRName leads to improved performance for common sparse linear algebra kernels across a variety of hardware platforms, including an Intel multi-core CPU, an NVIDIA GPU, an AMD Xilinx FPGA, and a simulated PIM device. \rev{Furthermore, we assess the performance of automatic format conversion (\S \ref{eval-conversion}) and compute kernel generation (\S \ref{eval-compute}) using \IRName. Our evaluation demonstrates that the programs generated by \IRName achieve performance that matches state-of-the-art approaches~\cite{bik2022compiler, chou2020automatic}, while \IRName offers broader coverage for handling a wider range of custom formats.}

\subsection{Experiment Setup}
\label{experiment-setup}
We obtain sparse matrices from various popular datasets, including SuiteSparse~\cite{suitesparse}, SNAP~\cite{snapnets}, and OGB~\cite{hu-ogb-neurips2020}, covering a rich mix of application domains. Additionally, we also collect a set of sparse weight tensors/matrices from a pruned Transformer model~\cite{sparse-weight}. Table \ref{table:ufl-dataset} summarizes all the sparse matrices used in our experiments. For improved readability, we use abbreviated names for these matrices in all the illustrations presented below.

\begin{table}[t]
\centering
\vspace{-5pt}
\caption{A summary of the matrices used for evaluation with the abbreviated names in parentheses.}
\vspace{-5pt}
\scriptsize
\setlength\tabcolsep{5pt} 
\resizebox{\textwidth}{!}{
\begin{tabular}{l c c c | l c c c}
\toprule
\thead{Matrix} & \thead{Shape} & \thead{Density} & \thead{Nonzero\\Diagonals} & \thead{Matrix} & \thead{Shape} & \thead{Density} & \thead{Nonzero\\Diagonals} \\
\midrule 
email-Eu-core (ee) & 1.01K $\times$ 1.01K & 2.5e-2 & 1.84K & 
ML\_Geer (ge) & 1.50M $\times$ 1.50M & 4.9e-5 & 9.35K\\
ss (ss) & 1.65M $\times$ 1.65M & 1.3e-5 &  1.68M &
ML\_Laplace (lp) & 377K $\times$ 377K & 1.9e-4 & 4.70K\\
Transport (tp) & 1.60M $\times$ 1.60M & 9.2e-6 &  15 &
rajat31 (ra) & 4.69M $\times$ 4.69M & 9.2e-7 & 5.05K\\
TSOPF\_RS\_b2383 (ts) & 38.1K $\times$ 38.1K & 1.1e-2 & 23.4K &
memchip (mc) & 2.71M $\times$ 2.71M & 2.0e-6 & 1.74M \\
vas\_stokes\_1M (vs) & 1.09M $\times$ 1.09M & 2.9e-5 & 1.67M &
crystm02 (cm) & 14.0K $\times$ 14.0K & 1.7e-3 & 27\\
cant (ct) & 62.5K $\times$ 62.5K & 1.0e-3 & 99 &
c8\_mat11 (c8) & 4.56K $\times$ 5.76K & 9.4e-2 & 10.3K\\
nemeth21 (nm) & 9.51K $\times$ 9.51K & 1.3e-2 & 169 &
heart1 (h1) & 3.56K $\times$ 3.56K & 1.1e-1 & 6.53K \\ 
bibd\_18\_9 (b9) & 153 $\times$ 48.6K & 2.4e-1 & 48.6K & 
cari (cr) & 400 $\times$ 1.20K & 3.2e-1 & 1.16K\\
transformer-0.5 (tf-0.5) & 512 $\times$ 33.3K & 5.0e-1 & 33.8K &
transformer-0.6 (tf-0.6) & 512 $\times$ 33.3K & 4.0e-1  & 33.8K\\
transformer-0.7 (tf-0.7) & 512 $\times$ 33.3K & 3.0e-1 & 33.8K &
transformer-0.8 (tf-0.8) & 512 $\times$ 33.3K & 2.0e-1 & 33.8K\\
transformer-0.9 (tf-0.9) & 512 $\times$ 33.3K & 1.0e-1 & 33.8K &
transformer-0.95 (tf-0.95) & 512 $\times$ 33.3K & 5.0e-2 & 33.7K\\
roadNet-PA (rp) & 1.09M $\times$ 1.09M & 1.3e-6 & 66.4K &
mouse\_gene (mg) & 45.1K $\times$ 45.1K & 7.1e-3 & 77.8K\\
google-plus (gp) & 108K $\times$ 108K & 1.2e-3 & 203K & 
pokec (pk) & 1.63M $\times$ 1.63M & 1.2e-5 & 2.28M\\
hollywood (hw) & 1.07M $\times$ 1.07M & 4.9e-5 & 2.08M & 
ogbl-ppa (op) & 576K $\times$ 576K & 1.3e-4 & 1.14M \\
LiveJournal (lj) & 4.85M $\times$ 4.85M & 2.9e-6 & 6.55M &
wikipedia-20051105 (wp) & 1.63M $\times$ 1.63M & 7.4e-6 & 2.86M\\
chem\_master1 (ch) & 40.4K $\times$ 40.4K & 1.2e-5 & 5 &
majorbasis (mj) & 160K $\times$ 160K & 6.8e-5 & 22 \\
shyy161 (sh) & 76.5K $\times$ 76.5K & 5.6e-5 & 7 &
Baumann (bm) & 112K $\times$ 112K & 5.9e-5 & 7 \\
wiki-Vote (wv) & 8.30K $\times$ 8.30K & 1.5e-3 & 11.4K & 
mario002 (m2) & 390K $\times$ 390K & 1.4e-5 & 507K \\
scircuit (sc) & 171K $\times$ 171K & 3.3e-5 & 159K &
p2pGnutella31 (pg) & 62.6K $\times$ 62.6K & 3.8e-5 & 53.2K\\
cage12 (ca) & 130K $\times$ 130K & 1.2e-4 & 75.5K &
filter3D (f3) & 106K $\times$ 106K & 2.4e-4 & 13.4K \\
ca-CondMat (cc) & 23.1K $\times$ 23.1K & 3.5e-4 & 40.2K &
poisson3Da (p3) & 13.5K $\times$ 13.5K & 1.9e-3 & 26.1K\\
bwm2000 (bw) & 2.00K $\times$ 2.00K & 2.0e-3 & 5 &
af23560 (af) & 23.6K $\times$ 23.6K & 8.7e-4 & 33 \\
cryg10000 (cg) & 10.0K $\times$ 10.0K & 5.0e-4 & 8 &
ex19 (ex) & 12.0K $\times$ 12.0K & 1.8e-3 & 185 \\
mycielskian12 (my) & 3.07K $\times$ 3.07K & 4.3e-2 & 6.12K &
ogbl-ddi (od) & 4.27K $\times$ 4.27K & 5.8e-2 & 8.50K \\
\bottomrule
\end{tabular}}
\label{table:ufl-dataset}
\vspace{-5pt}
\end{table}

All format conversion experiments are performed on a dual-socket 24C/24T Intel(R) Xeon(R) Gold 6248R CPU @ \SI{3.00}{GHz}. The compute kernels generated by \IRName are executed on multiple backends, including the same CPU, an NVIDIA GPU A6000, an AMD Xilinx FPGA, and a PIM-core simulator~\cite{devic2022pim}. We provide more details of the configurations for each experiment in the corresponding subsections. 
\subsection{Format Customization}
\label{eval-customization}
We present four case studies evaluating the format customization feature of \IRName. Our results demonstrate performance improvements on the CPU and the GPU using hybrid BDIA/CSR (\S \ref{eval-bdia}) and hybrid BELL/COO (\S \ref{eval-bell}) formats that are tailored to \rev{the non-zero} distribution patterns \rev{of matrices}. We showcase the versatility of \IRName by supporting the recently proposed Serpens format~\cite{song2022serpens}, which enables high-performance sparse processing on FPGAs (\S \ref{eval-serpens}). Furthermore, we demonstrate that \IRName can handle partitioned formats for PIM systems using C\textsuperscript{2}SR and CISR. Finally, we explore a more load-balanced version of CISR, referred to as CISR-plus, by leveraging the custom map functions and the \lstinline{Reorder} primitive in \IRName (\S \ref{eval-cisr}).

\subsubsection{Case Study: The Hybrid BDIA/CSR Format on CPUs}
\label{eval-bdia}
\begin{figure*}[ht]
    \vspace{-5pt}
    \begin{subfigure}{0.25\columnwidth}
        \includegraphics[scale=0.37]{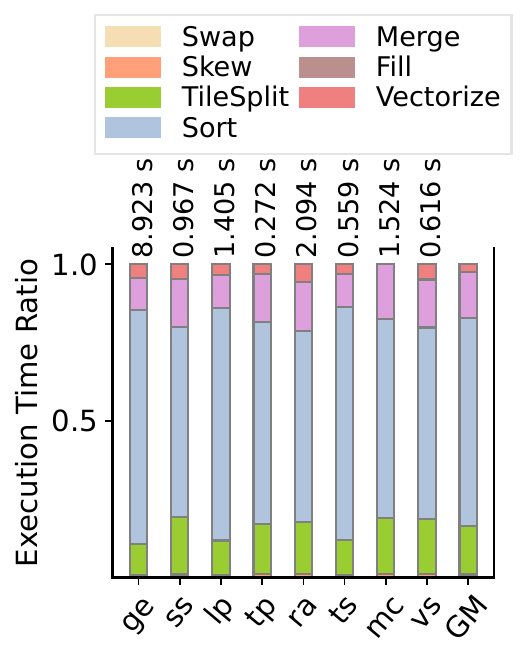}
    \end{subfigure}
    \begin{subfigure}{0.74\columnwidth}
        \centering
        \includegraphics[scale=0.32]{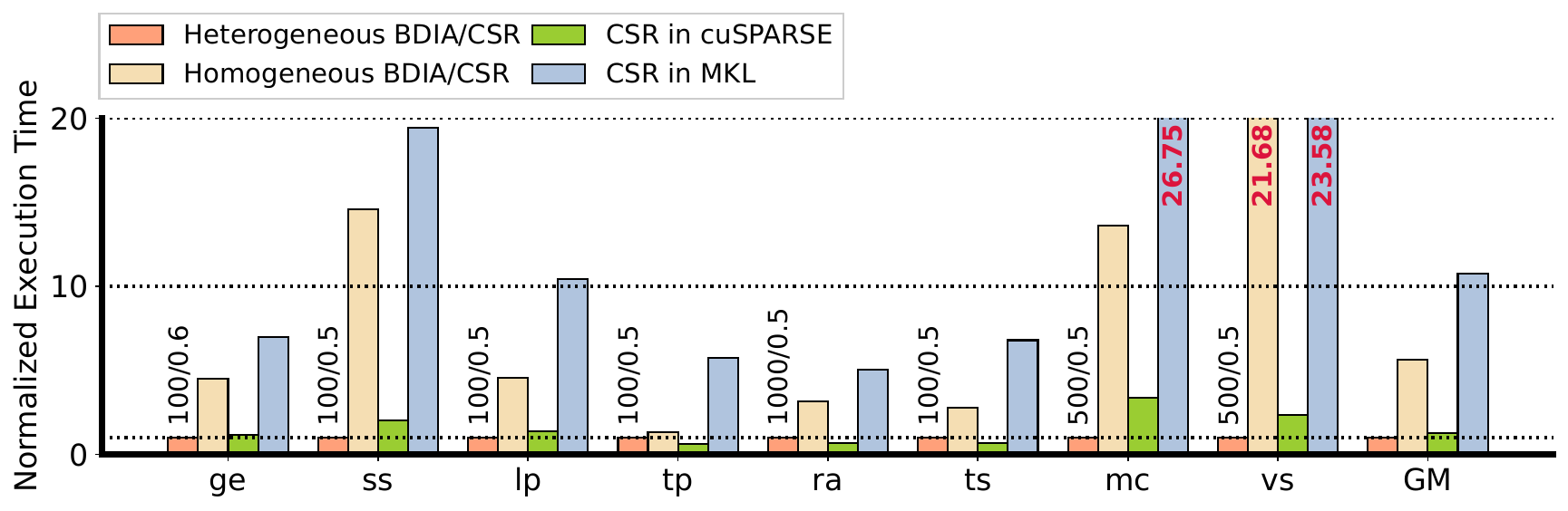}
    \end{subfigure}
    \vspace{-5pt}
    \caption{The case study of the hybrid BDIA/CSR format. The left figure shows the breakdown of the execution time for converting from COO to BDIA format. The right figure shows the performance comparison of the SpMV kernel across various configurations. The execution times are normalized to the heterogeneous configuration, and execution times over 20 are marked with red numbers inside bars.
    }
    \label{fig:hetero}
    \vspace{-5pt}
\end{figure*}

We generate the hybrid BDIA/CSR format using \IRName. 
Initially, the input matrix is decomposed into two sub-matrices, both represented in COO format. One sub-matrix is converted to the BDIA format, while the other is converted to the CSR format. The decomposition pattern is adjusted by tuning the blocking factor and the non-zero threshold of each diagonal, which can affect the performance of the sparse kernel. In this work, we manually select the blocking size and thresholds with the best performance. Automatically searching for the optimal decomposition factors will be left for future work. The block sizes and thresholds used for each dataset are presented above the result bars on the right side of Figure~\ref{fig:hetero}.

We profile the execution time breakdown of converting COO to BDIA on multiple datasets and summarize the results in the left side of Figure \ref{fig:hetero}. The \lstinline{Sort} operator is critical and accounts for 65\% of the total conversion time since it involves intensive memory reads and writes.

As shown in the right side of Figure \ref{fig:hetero}, we evaluate the SpMV kernel with the hybrid BDIA/CSR format (program in Figure \ref{fig:illustrate-unisparse-code}) \newrev{in single precision} on four hardware configurations, using a 48-core Intel Xeon Gold 6248R CPU at \SI{3.00}{GHz} and an NVIDIA RTX A6000 GPU. The SpMV kernel on the CPU is parallelized using OpenMP, and the kernel on the GPU leverages APIs provided by the cuSPARSE library. The heterogeneous configuration runs the SpMV kernel with one sub-matrix in BDIA on the CPU, and the other sub-matrix in CSR on the GPU simultaneously. The homogeneous configuration runs SpMV with both sub-matrices in BDIA and CSR formats on the CPU. Another two configurations run the SpMV kernel with only the CSR format implemented in cuSPARSE on the GPU, and in the Intel MKL library on the CPU. 
\newrev{For the MKL implementation of SpMV, we store the matrix in the CSR format (\lstinline{mkl_sparse_s_create_csr}), called the optimization function (\lstinline{mkl_sparse_optimize}) and the Inspector-Executor (IE) routine (\lstinline{mkl_sparse_s_mv}).}
The heterogeneous configuration yields 5.63$\times$, 1.28$\times$, and 10.73$\times$ speedup in Geomean over the SpMV kernel with the hybrid format only on the CPU (homogeneous configuration), using cuSPARSE on the GPU, and using MKL library on the CPU, respectively.

\subsubsection{Case Study: The Hybrid BELL/COO Format on GPUs}
\label{eval-bell}
\begin{figure*}[ht]
    \vspace{-5pt}
    \begin{subfigure}{0.25\columnwidth}
        \includegraphics[scale=0.37]{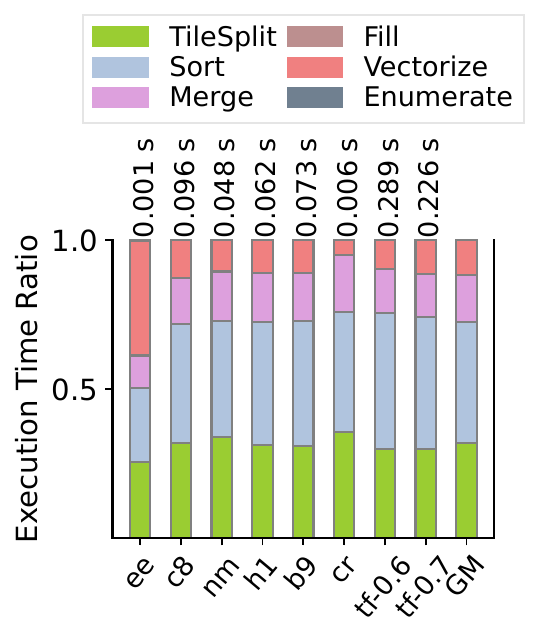}
    \end{subfigure}
    \begin{subfigure}{0.74\columnwidth}
        \centering
        \includegraphics[scale=0.32]{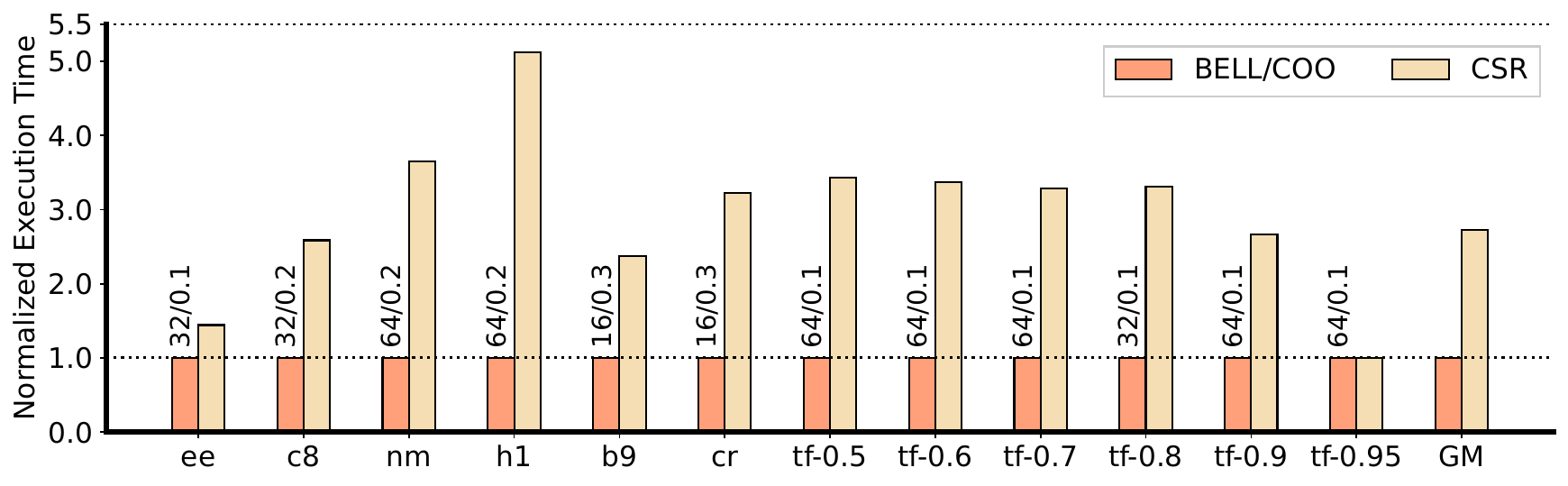}
    \end{subfigure}
    \vspace{-10pt}
    \caption{The case study of the hybrid BELL/COO format. The left figure shows the breakdown of execution time for converting from COO to the BELL format. The right figure shows the normalized run time of cuSPARSE SpMM using the BELL/COO decomposed by \IRName vs. CSR on NVIDIA A6000. Datasets transformer-50 to 95 are pruned weight matrices of Transformer~\cite{sparse-weight} with sparsity ranging from 50\% to 95\%.}
    \label{fig:bellpack}
    \vspace{-8pt}
\end{figure*}

\rev{The hybrid BELL/COO format is generated by decomposing the input matrix into two sub-matrices, one converting to the BELL format while the other remaining in the COO format. The time breakdown of the conversion from COO to BELL is profiled across multiple datasets, and the results are summarized on the left side of Figure~\ref{fig:bellpack}.
We manually tune the decomposition parameters by adjusting the block size and the non-zero threshold within each block, selecting the configuration that yields the best performance. These  decomposition parameters are indicated above each result bar on the right side of Figure~\ref{fig:bellpack}.}


We evaluate sparse matrix-matrix multiplication (SpMM) \newrev{in single precision} using the hybrid BELL/COO format and compare it with the one using only the CSR format. 
The compute kernel is deployed on an NVIDIA RTX A6000 GPU through APIs provided by the cuSPARSE library. 
Figure ~\ref{fig:bellpack} shows the normalized run time of the SpMM kernel using the CSR format vs. the BELL/COO format.  The hybrid BELL/COO format on the selected sparse matrices leads to a 2.7$\times$ Geomean speedup. 

\subsubsection{Case Study: The Serpens Format on FPGAs}
\label{eval-serpens}

\begin{figure*}[t]
\vspace{-5pt}
\begin{subfigure}{0.33\textwidth}
    \centering
    \includegraphics[scale=0.32]{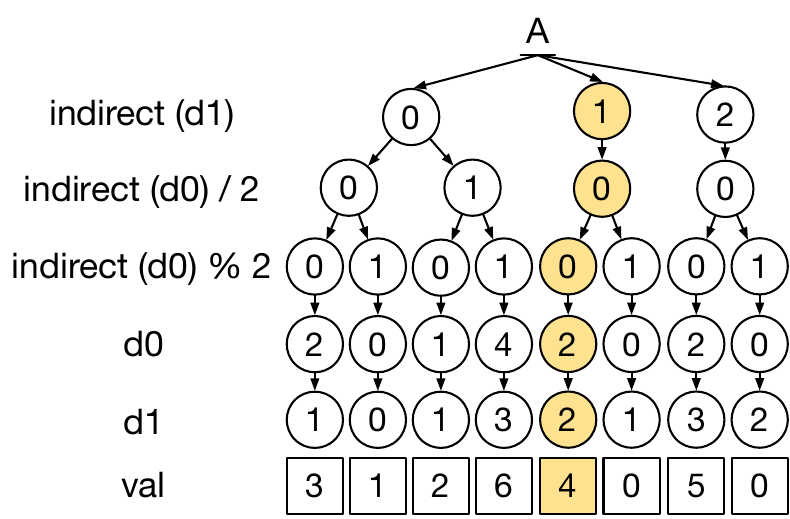}
    \caption{The data structure of Serpens}
    \label{tree-serpens}
\end{subfigure}
\begin{subfigure}{0.33\textwidth}
    \centering
    \includegraphics[scale=0.32]{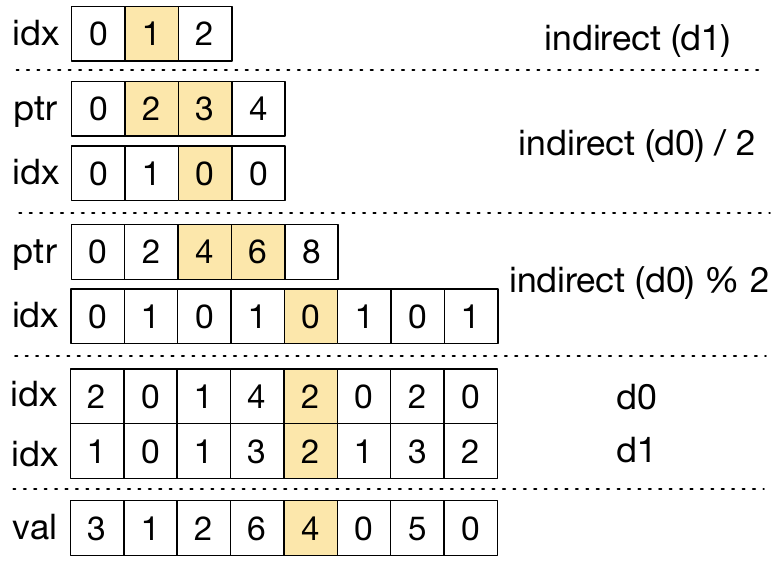}
    \caption{The data layout of Serpens}
    \label{layout-serpens}
\end{subfigure}
\begin{subfigure}{0.33\textwidth}
\centering
\centering
    \includegraphics[scale=0.35]{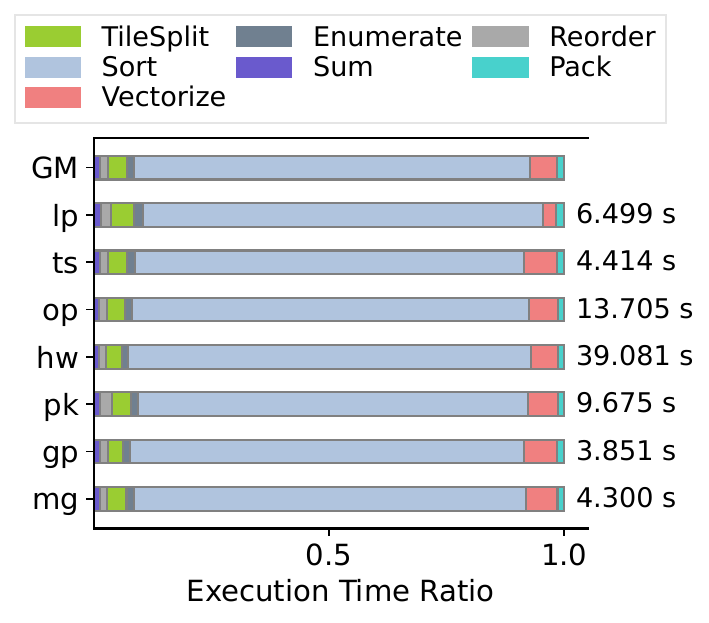}
    \caption{Breakdown of COO$\rightarrow$Serpens}
\label{serpens-breakdown}
\end{subfigure}
\vspace{-5pt}
\caption{The Serpens format of the matrix $A$ in Figure \ref{matrix}.}
\label{fig-serpens}
\vspace{-8pt}
\end{figure*}

We evaluate the sparse format proposed in the Serpens  accelerator~\cite{song2022serpens,song2022sextans} and demonstrate how \IRName can express and generate the Serpens format using indirect functions and query primitives that change the order of elements. As shown in Figures \ref{tree-serpens} and \ref{layout-serpens}, the Serpens format traverses elements in column order, preserving the dependency length between adjacent row elements with a few padding zeros. This allows the Serpens accelerator to achieve significant throughput improvement for the SpMV kernel.

\IRName automatically converts from COO to the Serpens format.
Despite the \lstinline{Sort} operator consuming a significant portion (84\% in Geomean) of the total time (Figure \ref{serpens-breakdown}), \IRName is able to generate this high-performance format in seconds, significantly improving the productivity of hardware developers and providing easier access to sparse acceleration for software developers. 
According to ~\cite{song2022serpens}, utilizing this custom format in Serpens results in 1.91$\times$ better throughput and 1.71$\times$ better energy efficiency on an AMD Xilinx Alveo U280 device compared to the prior state-of-the-art FPGA accelerator~\cite{hu2021graphlily}.

\subsubsection{Case Study: The C\textsuperscript{2}SR, CISR, and CISR-plus Formats on PIMs}
\label{eval-cisr}
\begin{figure*}[t]
\begin{subfigure}{0.33\textwidth}
    \centering
    \includegraphics[scale=0.32]{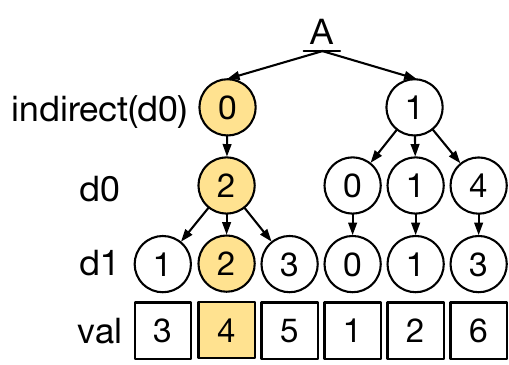}
    \caption{The data structure of CISR-plus}
    \label{tree-cisr-plus}
\end{subfigure}
\begin{subfigure}{0.33\textwidth}
    \centering
    \includegraphics[scale=0.32]{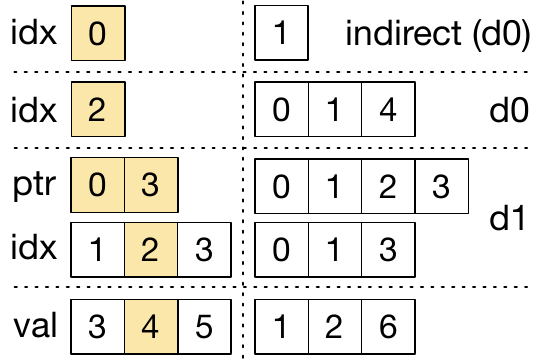}
    \caption{The data layout of CISR-plus}
    \label{layout-cisr-plus}
\end{subfigure}
\begin{subfigure}{0.33\textwidth}
\centering
\centering
    \includegraphics[scale=0.35]{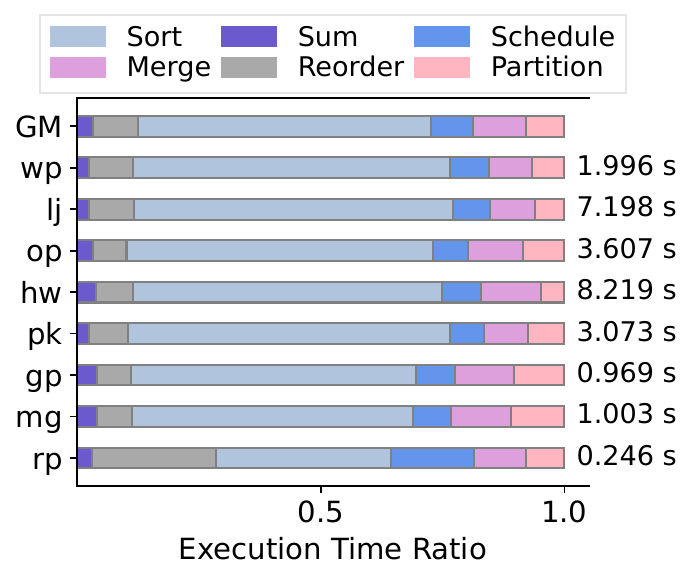}
    \caption{Breakdown of COO$\rightarrow$CISR-plus}
\label{cisr-plus-breakdown}
\end{subfigure}
\hfil
\begin{subfigure}{0.9\textwidth}
\centering
\begin{lstlisting}[
basicstyle={\scriptsize\ttfamily},
identifierstyle={\color{black}},
language={mlir},
tabsize=2,
numbersep=8pt,
numbers=left,
breaklines=true,
xleftmargin=2em, framexleftmargin=2em, aboveskip=1mm, belowskip=1mm,
escapeinside=''
]
#CISR-plus = #encoding<{ idx_map<(d0,d1)->(indirect(d0),d0,d1)>, mutation<merge(0,1), trim(1,2)>,
  indirect<{ sum(value) groupBy (d0,d1)->(d0) with value ne 0 -> 1 | otherwise -> 0 '\label{CISR_plus_indirect}'
             reorder(d0) traverseBy (d0,d1)->(d0)
             schedule(d0) traverseBy (d0,d1)->(d0/2) }>,
  layout<partition(0)> }>
\end{lstlisting}
\caption{The format encoding of CISR-plus.}
\label{encoding-cisr-plus}
\end{subfigure}
  

\vspace{-5pt}
\caption{The CISR-plus format of the matrix $A$ in Figure \ref{matrix}.}
\label{fig-cisr-plus}
\vspace{-8pt}
\end{figure*}

\begin{figure*}[ht]
\begin{subfigure}{0.9\textwidth}
    \centering
    \includegraphics[scale=0.3]{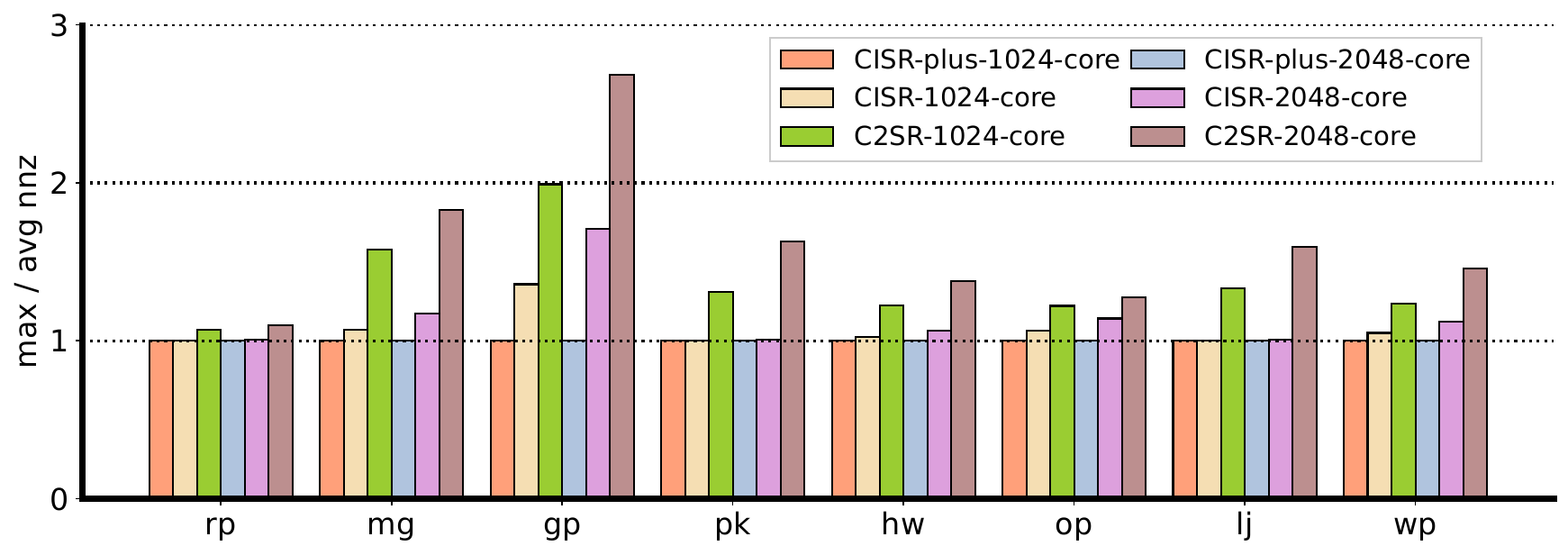}
    \caption{Load imbalance on different numbers of PIM cores.}
    \label{fig:imbalance}
\end{subfigure}
\begin{subfigure}{0.5\textwidth}
    \centering
    \includegraphics[scale=0.3]{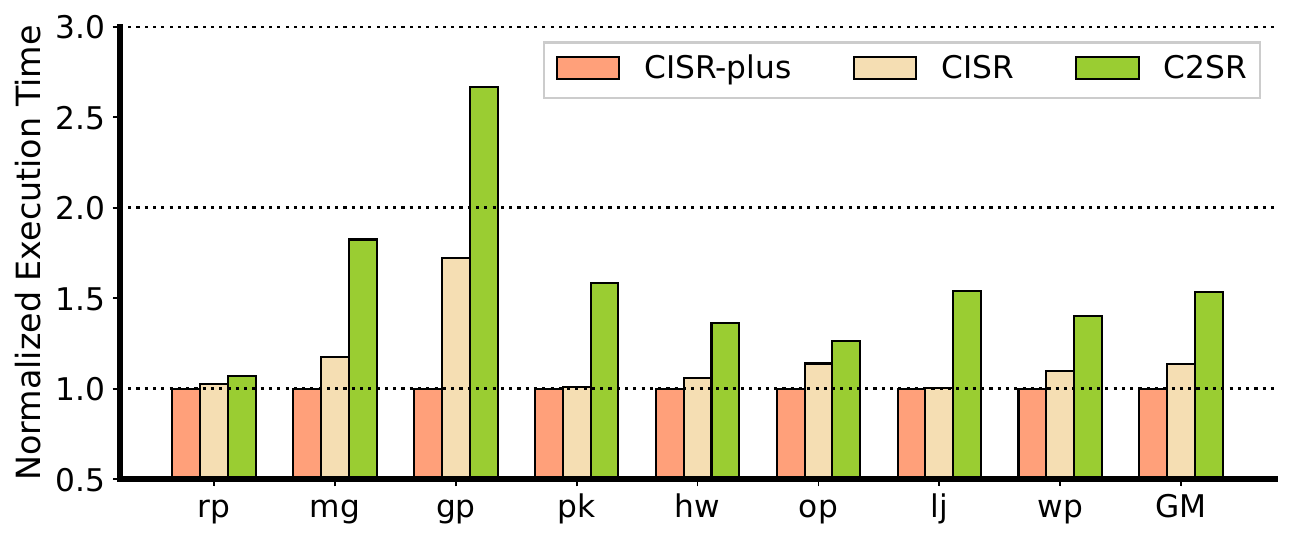}
    \caption{Normalized run time on 1024 PIM cores.}
    \label{fig:custcomp}
\end{subfigure}
\begin{subfigure}{0.48\textwidth}
    \centering
\vspace{5pt}
    \includegraphics[scale=0.35]{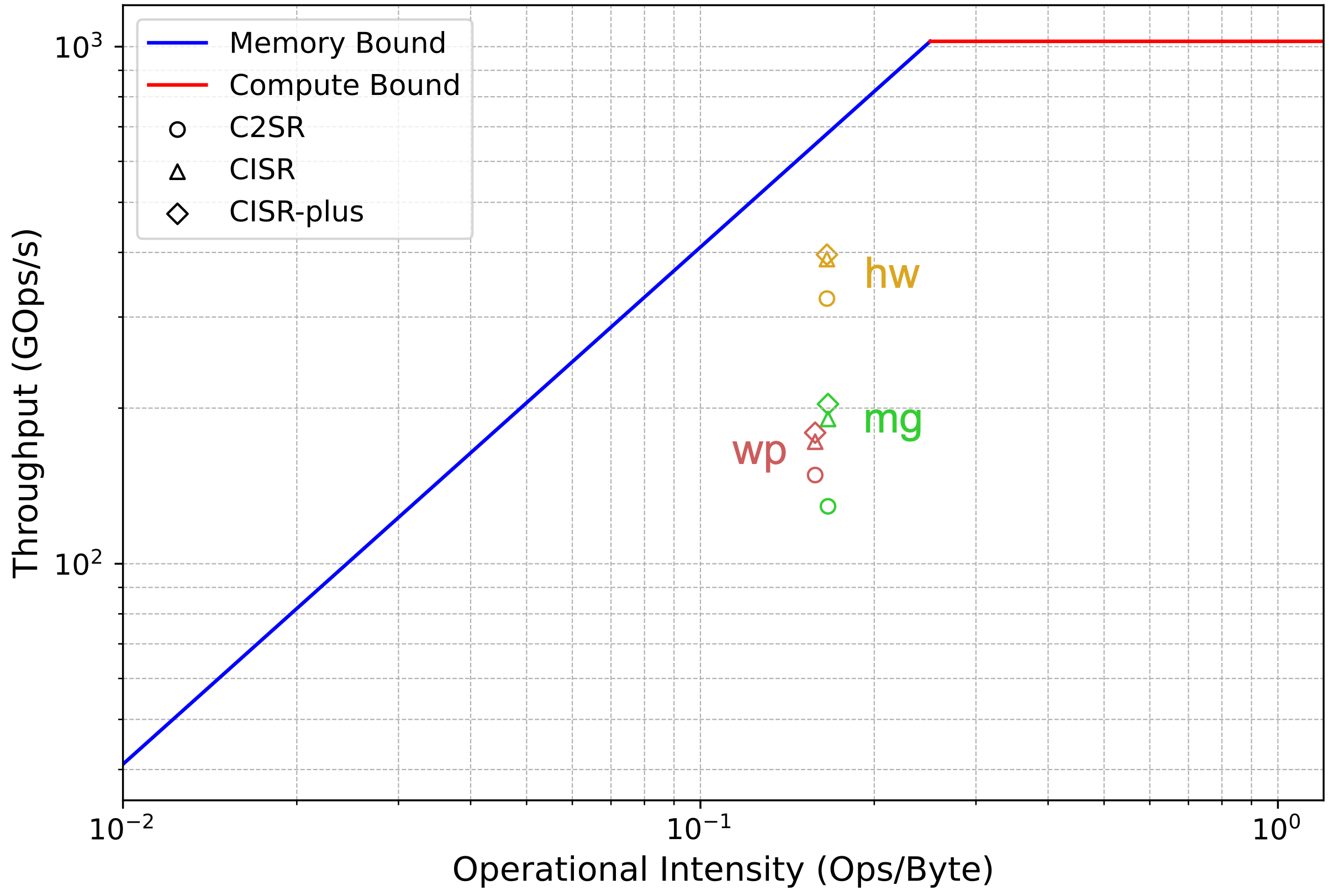}
    \caption{\newrev{Roofline analysis for 3 selected matrices on 1024 PIM cores.}}
\vspace{-5pt}
    \label{fig:roofline}
\end{subfigure}
\caption{The SpMV profiling results using CISR-plus, CISR, and C$^{2}$SR formats on the simulated PIM device.}
\vspace{-10pt}
\end{figure*}

We evaluate SpMV using the C\textsuperscript{2}SR and CISR formats on a simulated PIM device~\cite{devic2022pim} with 1024 and 2048 cores. 
Each PIM core has a copy of the input dense vector and computes a subset of the output vector in a lock-free execution pattern. 
Figure ~\ref{fig:imbalance} shows the maximum vs. the average number of non-zeros processed per core. As the number of cores increases, the load imbalance introduced by the C\textsuperscript{2}SR format gradually becomes a bottleneck, whereas the CISR format mitigates this issue. Figure~\ref{fig:custcomp} shows the normalized execution time of SpMV on 1024 PIM cores using the C\textsuperscript{2}SR vs. the CISR format. Compared with the C\textsuperscript{2}SR format, using the CISR format improves the performance by 1.35$\times$ in Geomean. 

Using \IRName, we can further explore an optimized version of CISR, called CISR-plus. CISR-plus schedules rows with the maximum number of non-zeros first, which leads to better load balance among PIM cores. The data structure and layout of CISR-plus are shown in Figures \ref{tree-cisr-plus} and \ref{layout-cisr-plus}. The \lstinline{reorder} primitive along with custom map functions enable us to express (Figure \ref{encoding-cisr-plus}) and generate the CISR-plus format using \IRName. The conversion time breakdown is shown in Figure \ref{cisr-plus-breakdown}. We can see that the \lstinline{Sort} operator still dominates the conversion time, while the \lstinline{Reorder} and \lstinline{Schedule} operators are relatively faster. \newrev{The improved load balance by using CISR-plus format is demonstrated in Figure \ref{fig:imbalance}.} We further evaluate the performance of CISR-plus by comparing it to CISR and C\textsuperscript{2}SR when running SpMV on 1024 PIM cores, as shown in Figure \ref{fig:custcomp}. Using CISR-plus leads to a 1.14$\times$ and 1.54$\times$ speedup in Geomean over CISR and C\textsuperscript{2}SR, respectively.

\newrev{Figure \ref{fig:roofline} presents a roofline analysis of employing C\textsuperscript{2}SR, CISR, and CISR-plus formats for 3 selected matrices using 1024 PIM cores. While all kernels are memory-bound, adopting the CISR-plus format brings the design nearer to the roofline by addressing the load imbalance issue among PIM cores, thereby increasing throughput.}

\subsubsection{\newrev{Case Study: Label Propagation on GPUs.}}
\label{eval-label-prop}
\newrev{
We evaluate the label propagation application using custom formats generated by \IRName and compare it against the traditional CSR format. Label propagation is a semi-supervised machine learning algorithm that iteratively propagates label information from previously labeled data points to unlabeled ones \cite{iscen2019label}. This algorithm operates through iterative matrix multiplications, where the right matrix, referred to as the label matrix, contains the label vectors for each data point (or node), and the left one is the adjacency matrix that represents the node connectivity. Initially, the label matrix is sparse, with only a few nodes having non-zero values in each label category. While employing sparse-sparse matrix multiplication (SpGEMM) for the initial stage of the algorithm may lead to better performance, as the density of the label matrix increases, it requires ad hoc experiments for each dataset to determine the density threshold to switch using SpMM for the rest of the computation. In our experiments, for simplicity, we use SpMM for the entire process of label propagation.
}

\newrev{
Table \ref{tab:label_prop} presents the evaluation results of the label propagation algorithm. The label propagation is implemented through 20 iterations of SpMM. Before computation, we normalize the sparse matrices to ensure each row sums up to 1, and we generate label matrices with a fixed column size of 1000. The hybrid BELL/COO format is generated by \IRName using 24 threads on an Intel Xeon Gold 6242 CPU at \SI{2.80}{GHz}. The compute kernels are executed on an NVIDIA RTX A6000 GPU through APIs provided by the cuSPARSE library. 
}

\newrev{
As presented in Table \ref{tab:label_prop}, the total execution time of the label propagation algorithm using the hybrid BELL/COO format includes a format preprocessing time and a kernel execution time. The preprocessing of the hybrid BELL/COO format involves intensive memory accesses to perform both matrix decomposition and format conversion, thus introducing a non-negligible overhead to overall performance. However, in applications involving multiple iterations of sparse linear algebra kernels, such as label propagation, the overhead from format preprocessing is amortized across iterations. This allows for an overall speedup by utilizing custom sparse formats. Our evaluation shows that using the hybrid BELL/COO format for the label propagation application leads to a 1.61$\times$ speedup in Geomean compared to the traditional CSR format.
}
\begin{table}[ht]
\centering
\caption{
\newrev{The case study of label propagation application. The numbers for each execution time (in \textit{milliseconds}) $\pm$ standard deviation are computed based on 20 independent runs.}}
\footnotesize
\setlength\tabcolsep{5pt} 
\lstset{%
tabsize=2,
columns=fullflexible,
basicstyle=\ttfamily\scriptsize,
breaklines=true,
breakatwhitespace,
escapeinside={(*}{*)},
alsoletter={\., \_},
morekeywords={tile\_division, tile\_union, enumerate, reorder, schedule, move\_lv, trim, fill, merge, split, vectorize, devectorize, pad, pack, partition}
}
\resizebox{0.8\textwidth}{!}{\newtablerev
\begin{tabular}{c|r|r|r|r|c}
\toprule
\multirow{2}{*}{\makecell{\thead{\textbf{Data}}}} & \multicolumn{1}{c|}{\multirow{2}{*}{\makecell{\thead{\textbf{CSR Format}}}}}  & \multicolumn{3}{c|}{\thead{\textbf{Hybrid BELL/COO Format}}} & \multirow{2}{*}{\makecell{\thead{\textbf{Speedup}\\\textbf{over CSR}}}} \\
\cmidrule{3-5} 
      &    & \makecell{Format Preprocessing} & Kernel Execution & \multicolumn{1}{c|}{Total} &    \\\midrule
  ee  & 11.969 $\pm$ 0.106  & 5.490 $\pm$ 0.644  &  4.460 $\pm$ 0.136  &   9.950 $\pm$ 0.658 & 1.20  \\
  c8  & 190.272 $\pm$ 1.149 & 63.022 $\pm$ 2.257 & 55.960 $\pm$ 0.246  & 118.983 $\pm$ 2.270 & 1.60  \\
  nm  & 96.842 $\pm$ 0.666  & 21.376 $\pm$ 2.665 & 49.809 $\pm$ 0.119  &  71.185 $\pm$ 2.667 & 1.36  \\
  h1  & 111.431 $\pm$ 0.589 & 27.172 $\pm$ 1.509 & 16.163 $\pm$ 0.125  &  43.335 $\pm$ 1.514 & 2.57  \\
  cr  & 11.764 $\pm$ 0.101  & 3.884 $\pm$ 0.146  &  3.661 $\pm$ 0.006  &   7.545 $\pm$ 0.146 & 1.56  \\
  my  & 32.629 $\pm$ 0.015  & 15.209 $\pm$ 0.555 & 11.162 $\pm$ 0.528  &  26.371 $\pm$ 0.766 & 1.24  \\
  od  & 246.337 $\pm$ 0.280 & 69.728 $\pm$ 3.687 & 55.450 $\pm$ 0.204  & 125.177 $\pm$ 3.693 & 1.97  \\
  wv  & 1464.949 $\pm$ 7.565 & 380.897 $\pm$ 10.194 & 458.716 $\pm$ 0.415 & 839.613 $\pm$ 10.202 & 1.74  \\\midrule
 Geomean &  \multicolumn{4}{c|}{}  & 1.61  \\
\bottomrule
\end{tabular}
}
\label{tab:label_prop}
\end{table}
\vspace{-8pt}

\subsection{Format Conversion}
\label{eval-conversion}
Compared to state-of-the-art compilers, \IRName \newrev{offers a wider coverage of supported formats in its automatic format conversion routines}. 
Table \ref{tab:conversion} lists 7 representative format conversion cases supported by \IRName. In the listed format conversion cases, both MLIR SparseTensor dialect~\cite{bik2022compiler} and TACO~\cite{chou2020automatic} lack stable support for DCSC $\rightarrow$ BCSR, CSB $\rightarrow$ DIA-variant, COO $\rightarrow$ C\textsuperscript{2}SR, and COO $\rightarrow$ CISR, while MLIR SparseTensor does not implement COO $\rightarrow$ DIA nor COO $\rightarrow$ ELL. Additionally, to the best of our knowledge, none of the prior sparse linear algebra compilers support specialized formats like CISR and Serpens. 

\newrev{Furthermore, \IRName demonstrates comparable performance in the given format conversion cases compared to two baseline compilers. Format conversion is memory-intensive, and both \IRName and TACO optimize their code to minimize memory accesses. However, differences in the implementations of code generation lead to performance variations. The format decoding in \IRName introduces performance overhead, which leads to inferior performance for CSR $\rightarrow$ CSC and COO $\rightarrow$ ELL conversions compared with TACO on small matrices (\lstinline{ch}, \lstinline{mj}, \lstinline{sh}, \lstinline{bm}) with high sparsity ($\sim$1.0e-5). We also notice that the transpose operation in MLIR SparseTensor utilizes an auxiliary class to enumerate values in any permutation order of tensor dimensions. While this approach offers functional versatility, it leads to increased time complexity and more memory accesses, resulting in slower performance when converting CSR to CSC.
}


\begin{table}[ht]
\centering
\caption{
Comparison of the actual execution times for format conversion programs generated by \IRName, \newrev{the MLIR SparseTensor dialect under the LLVM 15.0.0 release}, and \rev{the format conversion artifact of} TACO~\cite{github_chou2020automatic}. \newrev{The numbers for each execution time (in \textit{seconds}) $\pm$ standard deviation are computed based on 20 independent runs.}
}
\vspace{-5pt}
\footnotesize
\setlength\tabcolsep{5pt} 
\lstset{%
tabsize=2,
columns=fullflexible,
basicstyle=\ttfamily\scriptsize,
breaklines=true,
breakatwhitespace,
escapeinside={(*}{*)},
alsoletter={\., \_},
morekeywords={tile\_division, tile\_union, enumerate, reorder, schedule, move\_lv, trim, fill, merge, split, vectorize, devectorize, pad, pack, partition}
}
\resizebox{0.77\textwidth}{!}{\newtablerev
\begin{tabular}{c|ccc|c|c}
\toprule
\multirow{2}{*}{\makecell{\textbf{Data}}} & \multicolumn{3}{c|}{\textbf{CSR} $\rightarrow$ \textbf{CSC}} & \textbf{DCSC} $\rightarrow$ \textbf{BCSR} & \textbf{CSB} $\rightarrow$ \textbf{DIA-variant}  \\
\cmidrule{2-6} 
    & TACO & \makecell{SparseTensor} & \IRName & \IRName & \IRName  \\\midrule 
 wv & 0.001   $\pm$ 0.000 &  0.003   $\pm$ 0.000 & 0.001 $\pm$ 0.000 & 0.012 $\pm$ 0.001 & 0.067 $\pm$ 0.001  \\
 ee & 0.000   $\pm$ 0.000 &  0.001   $\pm$ 0.000 & 0.000 $\pm$ 0.000 & 0.003 $\pm$ 0.000 & 0.004 $\pm$ 0.000  \\
 nm & 0.005   $\pm$ 0.000 &  0.031   $\pm$ 0.001 & 0.002 $\pm$ 0.000 & 0.039 $\pm$ 0.000 & 0.032 $\pm$ 0.001  \\
 cm & 0.001   $\pm$ 0.000 &  0.006   $\pm$ 0.000 & 0.001 $\pm$ 0.000 & 0.012 $\pm$ 0.001 & 0.007 $\pm$ 0.000  \\
 ct & 0.018   $\pm$ 0.000 &  0.078   $\pm$ 0.004 & 0.009 $\pm$ 0.000 & 0.148 $\pm$ 0.001 & 0.184 $\pm$ 0.002  \\
 lp & 0.191   $\pm$ 0.004 &  0.659   $\pm$ 0.005 & 0.193 $\pm$ 0.003 & 2.554 $\pm$ 0.015 & 4.509 $\pm$ 0.031  \\
 tp & 0.168   $\pm$ 0.004 &  0.562   $\pm$ 0.004 & 0.139 $\pm$ 0.002 & 2.249 $\pm$ 0.010 & 6.157 $\pm$ 0.038  \\
 ts & 0.162   $\pm$ 0.002 &  0.462   $\pm$ 0.006 & 0.168 $\pm$ 0.001 & 1.733 $\pm$ 0.009 & 2.509 $\pm$ 0.017  \\
 ch & 0.001   $\pm$ 0.000 &  0.006   $\pm$ 0.000 & 0.001 $\pm$ 0.000 & 0.015 $\pm$ 0.001 & 0.009 $\pm$ 0.000  \\
 mj & 0.006   $\pm$ 0.000 &  0.043   $\pm$ 0.001 & 0.008 $\pm$ 0.000 & 0.134 $\pm$ 0.001 & 0.178 $\pm$ 0.002  \\
 sh & 0.001   $\pm$ 0.000 &  0.009   $\pm$ 0.001 & 0.001 $\pm$ 0.000 & 0.023 $\pm$ 0.000 & 0.018 $\pm$ 0.000  \\
 bm & 0.002   $\pm$ 0.000 &  0.018   $\pm$ 0.000 & 0.003 $\pm$ 0.000 & 0.058 $\pm$ 0.001 & 0.063 $\pm$ 0.001  \\\midrule
 \makecell{Geomean\\Speedup} &   1              &  0.20             & 1.13              &    1                             &  1    \\
\bottomrule
\end{tabular}
}
\begin{adjustwidth}{1.6cm}{}
\resizebox{0.82\textwidth}{!}{\newtablerev
\begin{tabular}{c|cc|cc|c|c}
\toprule
\multirow{2}{*}{\makecell{\textbf{Data}}} & \multicolumn{2}{c|}{\textbf{COO} $\rightarrow$ \textbf{ELL}} & \multicolumn{2}{c|}{\textbf{COO} $\rightarrow$ \textbf{DIA}} & \textbf{COO} $\rightarrow$ \textbf{C\textsuperscript{2}SR} & \textbf{COO} $\rightarrow$ \textbf{CISR}\\
\cmidrule{2-7} 
    & TACO & \IRName & TACO & \IRName & \IRName & \IRName \\\midrule
 wv & 0.018 $\pm$ 0.000 & 0.014 $\pm$ 0.001 & 0.092 $\pm$ 0.001 & 0.065 $\pm$ 0.004 & 0.003 $\pm$ 0.000 & 0.003 $\pm$ 0.000 \\
 ee & 0.002 $\pm$ 0.000 & 0.001 $\pm$ 0.000 & 0.005 $\pm$ 0.000 & 0.003 $\pm$ 0.000 & 0.001 $\pm$ 0.000 & 0.001 $\pm$ 0.000 \\
 nm & 0.011 $\pm$ 0.000 & 0.006 $\pm$ 0.000 & 0.008 $\pm$ 0.000 & 0.003 $\pm$ 0.000 & 0.017 $\pm$ 0.000 & 0.017 $\pm$ 0.000 \\
 cm & 0.002 $\pm$ 0.000 & 0.002 $\pm$ 0.000 & 0.001 $\pm$ 0.000 & 0.001 $\pm$ 0.000 & 0.005 $\pm$ 0.000 & 0.005 $\pm$ 0.000 \\
 ct & 0.048 $\pm$ 0.000 & 0.023 $\pm$ 0.000 & 0.031 $\pm$ 0.000 & 0.013 $\pm$ 0.000 & 0.071 $\pm$ 0.001 & 0.077 $\pm$ 0.001 \\
 lp & 0.289 $\pm$ 0.006 & 0.298 $\pm$ 0.006 & 0.433 $\pm$ 0.004 & 0.352 $\pm$ 0.008 & 1.069 $\pm$ 0.004 & 1.080 $\pm$ 0.021 \\
 tp & 0.208 $\pm$ 0.005 & 0.220 $\pm$ 0.002 & 0.133 $\pm$ 0.003 & 0.127 $\pm$ 0.002 & 0.929 $\pm$ 0.009 & 1.303 $\pm$ 0.062 \\
 ts & 0.274 $\pm$ 0.003 & 0.249 $\pm$ 0.004 & 0.192 $\pm$ 0.002 & 0.163 $\pm$ 0.002 & 0.541 $\pm$ 0.005 & 0.508 $\pm$ 0.045 \\
 ch & 0.001 $\pm$ 0.000 & 0.002 $\pm$ 0.000 & 0.001 $\pm$ 0.000 & 0.001 $\pm$ 0.000 & 0.006 $\pm$ 0.001 & 0.007 $\pm$ 0.001 \\
 mj & 0.010 $\pm$ 0.000 & 0.017 $\pm$ 0.000 & 0.010 $\pm$ 0.000 & 0.008 $\pm$ 0.000 & 0.060 $\pm$ 0.001 & 0.072 $\pm$ 0.001 \\
 sh & 0.002 $\pm$ 0.000 & 0.003 $\pm$ 0.000 & 0.002 $\pm$ 0.000 & 0.001 $\pm$ 0.000 & 0.011 $\pm$ 0.000 & 0.013 $\pm$ 0.000 \\
 bm & 0.004 $\pm$ 0.000 & 0.007 $\pm$ 0.000 & 0.003 $\pm$ 0.000 & 0.004 $\pm$ 0.000 & 0.026 $\pm$ 0.000 & 0.031 $\pm$ 0.000 \\\midrule
 \makecell{Geomean\\Speedup} &   1              & 1.01  &     1                 & 1.37             &    1                 &     1                \\
\bottomrule
\end{tabular}
}
\end{adjustwidth}
\label{tab:conversion}
\vspace{-5pt}
\end{table}

\subsection{Compute Kernel Generation}
\label{eval-compute}
Table \ref{tab:codegen} provides a performance comparison of SpMM and SpGEMM kernels across various formats generated by \IRName, MLIR SparseTensor, and TACO. \newrev{The SpMM kernel performs matrix multiplication between a sparse matrix in the dataset and a synthetic dense matrix with a fixed column size of 1000. The SpGEMM kernel conducts matrix multiplication between a sparse matrix in the dataset and itself.}
These kernels operate with double precision and execute in a single-threaded environment on an Intel Xeon Gold 6248R CPU at \SI{3.00}{GHz}. 

\newrev{\IRName leverages the kernel generation passes from the MLIR SparseTensor dialect for compute kernels including CSR SpMM, DCSC SpMM, CSR$\times$CSR$\rightarrow$CSR SpGEMM, and CSC$\times$CSC$\rightarrow$CSC SpGEMM. However, \IRName and MLIR SparseTensor differ in the implementation of the tensor storage. The tensor initialization function of MLIR SparseTensor is slightly more complicated. Therefore, the compute kernels generated by MLIR SparseTensor are slightly slower in some cases but mostly on par with \IRName. }

\newrev{The compute kernels generated by TACO show slower performance due to its different kernel generation strategies. For instance, in the case of CSR$\times$CSR$\rightarrow$CSR SpGEMM, all three compilers generate the row-wise product implementation, while the TACO-generated kernel involves an additional sorting of a partial sum index buffer, which makes it slower. We exclude the TACO column for the CSC$\times$CSC$\rightarrow$CSC SpGEMM kernel because 
the kernel generated by TACO produces incorrect outputs according to our experiments.}

\newrev{Furthermore, \IRName generates the SpMM kernel with the DIA-variant format using the kernel generation method described in Section \ref{compute-codegen}, whereas MLIR SparseTensor and TACO lack support for this format. From Table \ref{tab:codegen}, we observe that the performance of SpMM using the DIA-variant format is inferior compared to the CSR or DCSC format on the common datasets \lstinline{ee}, \lstinline{ch}, \lstinline{sh}, \lstinline{mj}, and \lstinline{bm}. This performance gap is primarily due to the extra computation required for zero paddings along the diagonals in the DIA-variant format during single-threaded execution. 
As presented in Section \ref{eval-bdia} and \ref{eval-bell}, formats with zero paddings exhibit better performance in a multi-threaded setting where computation among the padded dimensions can be parallelized.}


\begin{table}[ht]
\centering
\caption{
Comparison of the actual execution times for compute kernels generated by \IRName, \newrev{the MLIR SparseTensor dialect under the LLVM 15.0.0 release}, and \newrev{the main branch of TACO~\cite{github_taco}}. \newrev{The numbers for each execution time (in \textit{seconds}) $\pm$ standard deviation are computed based on 20 independent runs.}}
\footnotesize
\setlength\tabcolsep{5pt} 
\lstset{%
tabsize=2,
columns=fullflexible,
basicstyle=\ttfamily\scriptsize,
breaklines=true,
breakatwhitespace,
escapeinside={(*}{*)},
alsoletter={\., \_},
morekeywords={tile\_division, tile\_union, enumerate, reorder, schedule, move\_lv, trim, fill, merge, split, vectorize, devectorize, pad, pack, partition}
}
\resizebox{0.95\textwidth}{!}{\newtablerev
\begin{tabular}{c|rcc|rcc|c|r}
\toprule
\multirow{2}{*}{\makecell{\thead{\textbf{Data}}}} & \multicolumn{3}{c|}{\thead{\textbf{CSR SpMM}}} & \multicolumn{3}{c|}{\thead{\textbf{DCSC SpMM}}}  & \multirow{2}{*}{\makecell{\thead{\textbf{Data}}}} & \multicolumn{1}{c}{\thead{\textbf{DIA-variant}\\\textbf{SpMM}}} \\
\cmidrule{2-7} \cmidrule{9-9}
    & \multicolumn{1}{c}{TACO} & \makecell{SparseTensor} & \IRName & \multicolumn{1}{c}{TACO} & \makecell{SparseTensor} & \IRName &  & \multicolumn{1}{c}{\IRName}  \\\midrule
   m2    & 9.682  $\pm$ 0.109 & 1.937 $\pm$ 0.309 & 1.847 $\pm$ 0.038 & 9.608  $\pm$ 0.059 & 1.889 $\pm$ 0.086 & 1.857 $\pm$ 0.018 & bw &  0.078 $\pm$ 0.001  \\
   sc    & 4.344  $\pm$ 0.046 & 0.727 $\pm$ 0.005 & 0.734 $\pm$ 0.042 & 4.410  $\pm$ 0.314 & 0.739 $\pm$ 0.034 & 0.730 $\pm$ 0.007 & af &  8.136 $\pm$ 0.135  \\
   pg    & 0.877  $\pm$ 0.009 & 0.154 $\pm$ 0.002 & 0.153 $\pm$ 0.006 & 0.875  $\pm$ 0.011 & 0.151 $\pm$ 0.006 & 0.150 $\pm$ 0.002 & cg &  0.528 $\pm$ 0.012  \\
   ca    & 8.162  $\pm$ 0.495 & 1.534 $\pm$ 0.020 & 1.498 $\pm$ 0.030 & 7.983  $\pm$ 0.080 & 1.567 $\pm$ 0.062 & 1.528 $\pm$ 0.027 & ex & 22.642 $\pm$ 0.323  \\
   f3    & 10.327 $\pm$ 0.137 & 1.749 $\pm$ 0.139 & 1.702 $\pm$ 0.059 & 10.247 $\pm$ 0.079 & 1.765 $\pm$ 0.055 & 1.729 $\pm$ 0.038 & cm &  2.026 $\pm$ 0.027 \\
   cc    & 0.818  $\pm$ 0.011 & 0.185 $\pm$ 0.004 & 0.187 $\pm$ 0.009 & 0.813  $\pm$ 0.008 & 0.188 $\pm$ 0.006 & 0.187 $\pm$ 0.003 & ct & 32.896 $\pm$ 0.460  \\
   p3    & 1.364  $\pm$ 0.020 & 0.298 $\pm$ 0.007 & 0.302 $\pm$ 0.021 & 1.353  $\pm$ 0.013 & 0.302 $\pm$ 0.006 & 0.302 $\pm$ 0.006 & nm &  7.841 $\pm$ 0.133  \\
   ee    & 0.098  $\pm$ 0.002 & 0.014 $\pm$ 0.001 & 0.014 $\pm$ 0.001 & 0.098  $\pm$ 0.002 & 0.014 $\pm$ 0.001 & 0.014 $\pm$ 0.001 & ee &  3.806 $\pm$ 0.044  \\
   ch    & 0.918  $\pm$ 0.007 & 0.132 $\pm$ 0.007 & 0.136 $\pm$ 0.009 & 0.917  $\pm$ 0.007 & 0.136 $\pm$ 0.008 & 0.135 $\pm$ 0.008 & ch &  2.179 $\pm$ 0.084 \\
   sh    & 1.561  $\pm$ 0.008 & 0.223 $\pm$ 0.015 & 0.223 $\pm$ 0.016 & 1.566  $\pm$ 0.012 & 0.230 $\pm$ 0.015 & 0.222 $\pm$ 0.002 & sh &  5.821 $\pm$ 0.202  \\
   mj    & 6.997  $\pm$ 0.049 & 1.165 $\pm$ 0.196 & 1.133 $\pm$ 0.124 & 6.993  $\pm$ 0.051 & 1.101 $\pm$ 0.026 & 1.125 $\pm$ 0.024 & mj & 30.623 $\pm$ 0.472  \\
   bm    & 3.266  $\pm$ 0.011 & 0.533 $\pm$ 0.031 & 0.542 $\pm$ 0.049 & 3.274  $\pm$ 0.024 & 0.546 $\pm$ 0.043 & 0.529 $\pm$ 0.005 & bm &  8.672 $\pm$ 0.235 \\\midrule
 \makecell{Geomean\\Speedup} & \multicolumn{1}{c}{1}    & 5.77             & 5.81        & \multicolumn{1}{c}{1}        & 5.72             & 5.81   &    & \multicolumn{1}{c}{1}  \\
\bottomrule
\end{tabular}
}
\begin{adjustwidth}{0.33cm}{}
\resizebox{0.66\textwidth}{!}{\newtablerev
\begin{tabular}{c|ccc|cc}
\toprule
\multirow{2}{*}{\makecell{\thead{\textbf{Data}}}} & \multicolumn{3}{c|}{\thead{\textbf{CSR}{$\times$}\textbf{CSR}$\rightarrow$\textbf{CSR}\\\textbf{SpGEMM}}} & \multicolumn{2}{c}{\thead{\textbf{CSC}{$\times$}\textbf{CSC}$\rightarrow$\textbf{CSC}\\\textbf{SpGEMM}}} \\
\cmidrule{2-6}
    & TACO  & \makecell{SparseTensor}  & \IRName & \makecell{SparseTensor} & \IRName  \\\midrule
   m2    & 0.610 $\pm$ 0.010 & 0.390 $\pm$ 0.004 & 0.305 $\pm$ 0.006 & 0.388 $\pm$ 0.005 & 0.309 $\pm$ 0.006 \\
   sc    & 0.434 $\pm$ 0.003 & 0.245 $\pm$ 0.002 & 0.242 $\pm$ 0.003 & 0.244 $\pm$ 0.003 & 0.244 $\pm$ 0.008 \\
   pg    & 0.050 $\pm$ 0.002 & 0.032 $\pm$ 0.001 & 0.028 $\pm$ 0.001 & 0.031 $\pm$ 0.001 & 0.025 $\pm$ 0.001 \\
   ca    & 1.384 $\pm$ 0.010 & 0.725 $\pm$ 0.007 & 0.657 $\pm$ 0.014 & 0.720 $\pm$ 0.012 & 0.654 $\pm$ 0.006 \\
   f3    & 2.041 $\pm$ 0.017 & 1.178 $\pm$ 0.017 & 1.058 $\pm$ 0.008 & 1.166 $\pm$ 0.010 & 1.062 $\pm$ 0.008 \\
   cc    & 0.268 $\pm$ 0.004 & 0.146 $\pm$ 0.002 & 0.137 $\pm$ 0.001 & 0.143 $\pm$ 0.002 & 0.137 $\pm$ 0.002 \\
   p3    & 0.385 $\pm$ 0.005 & 0.191 $\pm$ 0.003 & 0.179 $\pm$ 0.003 & 0.189 $\pm$ 0.002 & 0.178 $\pm$ 0.002 \\
   ee    & 0.055 $\pm$ 0.005 & 0.021 $\pm$ 0.002 & 0.020 $\pm$ 0.001 & 0.021 $\pm$ 0.001 & 0.019 $\pm$ 0.001 \\
   ch    & 0.034 $\pm$ 0.001 & 0.012 $\pm$ 0.001 & 0.012 $\pm$ 0.001 & 0.012 $\pm$ 0.000 & 0.013 $\pm$ 0.000 \\
   sh    & 0.047 $\pm$ 0.001 & 0.020 $\pm$ 0.002 & 0.020 $\pm$ 0.001 & 0.019 $\pm$ 0.000 & 0.021 $\pm$ 0.004 \\
   mj    & 0.504 $\pm$ 0.006 & 0.251 $\pm$ 0.022 & 0.219 $\pm$ 0.015 & 0.242 $\pm$ 0.013 & 0.217 $\pm$ 0.001 \\
   bm    & 0.154 $\pm$ 0.001 & 0.079 $\pm$ 0.006 & 0.079 $\pm$ 0.007 & 0.077 $\pm$ 0.004 & 0.077 $\pm$ 0.005 \\\midrule
\makecell{Geomean\\Speedup}  & 1                     & 1.98             & 2.15             & 1                     & 1.06             \\
\bottomrule
\end{tabular}
}
\end{adjustwidth}
\label{tab:codegen}
\end{table}
\vspace{-5pt}


\section{Conclusion}
We present \IRName, an intermediate language that describes sparse tensor formats using decoupled data structures and data layouts. Our approach formulates the data structures of a sparse format virtually as a \Tree, described using an \Map and a set of structure mutation primitives. To enhance the expressibility of sparse data structures, we introduce query primitives that allow custom \Map functions. Data layouts are determined using layout primitives that enable switching from SoA to AoS layout and partitioning a tensor. With the well-formulated format description, the \IRName compiler automates format customization, as well as code generation for format conversion and compute kernels. Our work facilitates research into efficient formats for various types of tensors, expediting the development of fast sparse tensor algebra in diverse domains, including machine learning, scientific computing, and data analytics.

\section*{Data-Availability Statement}
The software supporting this paper is maintained publicly on GitHub~\cite{Liu_UniSparse_An_Intermediate_2024}. The version submitted to the OOPSLA' 24 Artifact Evaluation Committee (AEC) is permanently archived on Zenodo~\cite{jie_liu_2024_10464500}.
\begin{acks}                            

  This work was supported in part by ACE, one of the seven centers in JUMP 2.0, a Semiconductor Research Corporation (SRC) program sponsored by DARPA, NSF Awards \#1909661, \#2019306, \#2118709 and \#2212371, and by AFRL and DARPA under agreement FA8650-18-2-7863.
\end{acks}

\bibliography{ref}

\appendix


\end{document}